\pdfoutput=1

\documentclass[conference]{IEEEtran}
\usepackage{graphicx}

\usepackage{url}

\usepackage[breaklinks]{hyperref}

\usepackage{pbox}
\usepackage{xcolor}

\usepackage{floatrow}
\usepackage{subfig}

\usepackage{cleveref}
\usepackage{flushend}
\usepackage{tablefootnote}
\usepackage[flushleft]{threeparttable}
\usepackage{float}
\usepackage{multicol}

\def\BibTeX{{\rm B\kern-.05em{\sc i\kern-.025em b}\kern-.08em
    T\kern-.1667em\lower.7ex\hbox{E}\kern-.125emX}}

\DeclareGraphicsExtensions{.pdf,.png,.jpg}
\graphicspath{{./ImportedFigures/}}

\ifCLASSINFOpdf
\else
\fi
\newif\ifdraft
\drafttrue
\ifdraft
\usepackage{xcolor}
\usepackage{array}
\usepackage{makecell}

\definecolor{ocolor}{rgb}{1,0,0.4}
\else
\newcommand{\aanote}[1]{}
\newcommand{\alnote}[1]{}
\newcommand{\ednote}[1]{}
\newcommand{\kknote}[1]{}
\newcommand{\dungnote}[1]{}
\newcommand{\sanote}[1]{}
\fi

\begin{document}

\title{Synthetic Image Data for Deep Learning}

\author{
\IEEEauthorblockN{Jason W. Anderson}
\IEEEauthorblockA{\textit{BMW IT Research Center} \\
Greenville, SC \\
\small jason.anderson@bmwgroup.com}
\and
\IEEEauthorblockN{Marcin Ziolkowski}
\IEEEauthorblockA{\textit{BMW IT Research Center} \\
Greenville, SC \\
\small marcin.ziolkowski@bmwgroup.com
}
\and
\IEEEauthorblockN{Ken Kennedy}
\IEEEauthorblockA{\textit{BMW IT Research Center} \\
Greenville, SC \\
\small ken.kennedy@bmwgroup.com}
\and
\IEEEauthorblockN{Amy W. Apon}
\IEEEauthorblockA{\textit{School of Computing} \\
\textit{Clemson University}\\
Clemson, USA \\
\small aapon@clemson.edu}
}

\maketitle


\begin{abstract}
Realistic synthetic image data rendered from 3D models can be used to augment image sets and train image classification semantic segmentation models.
In this work, we explore how high quality physically-based rendering and domain randomization can efficiently create a large synthetic dataset based on production 3D CAD models of a real vehicle.
We use this dataset to quantify the effectiveness of synthetic augmentation using U-net and Double-U-net models.
We found that, for this domain, synthetic images were an effective technique for augmenting limited sets of real training data.  
We observed that models trained on purely synthetic images had a very low mean prediction IoU on real validation images.
We also observed that adding even very small amounts of real images to a synthetic dataset greatly improved accuracy, and that models trained on datasets augmented with synthetic images were more accurate than those trained on real images alone.

Finally, we found that in use cases that benefit from incremental training or model specialization, pretraining a base model on synthetic images provided a sizeable reduction in the training cost of transfer learning, allowing up to 90\% of the model training to be front-loaded.


\end{abstract}





\section{Introduction}
\label{sec:introduction}
In the field of image classification and segmentation with deep learning systems, access to sets of labelled training images with sufficient quantity and quality can be a formidable barrier to training an accurate model.  Collecting, segmenting, and labelling high quality images can be prohibitively expensive both in time and monetary cost.  In some cases, the barrier can be lowered by pretraining a model with a generic dataset such as ImageNet\cite{deng_imagenet_2009} and then fine-tuned on a smaller set of images more directly related to the project goals.  However, depending on the specificity requirements for the final model, a generalized dataset may not be useful.

A common alternative to vast quantities of readily available general images and costly task-specific images is synthetic image generation, where a 3D computer model of a scene relevant to the deep learning model is rendered to an image, segmented and/or classified, and then used to augment the training data available to the model.  Synthetic data has been used successfully in a growing body of research, in many cases reducing the overall cost of training a model.

Advantages to using synthetic images are not limited to overcoming the time and safety constraints of capturing and annotating real images.
3D modeling systems are very flexible -- scenes and assets can be changed and re-rendered with a cost likely far less than the real world equivalent.
For example, in the use cases presented in this work, the cost of changing the vehicle CAD model to a brand new vehicle or a new model year and then generating a new training set is far less than that of acquiring new real world examples, especially when the goal is to have a working detection system before the model enters production.
The costs of developing a synthetic image generation pipeline specific to a model's goals can be further recuperated in cases where similar images can be used to train other models, potentially requiring only minor alterations to the generator.   

While the body of work around using synthetic images in deep learning models has become broadened in recent years, we have found little exploration of using synthetic images to pretrain a multistage segmentation model such as the recently proposed DoubleU-net which has been shown to be highly accurate in some applications.  Our motivations in this work are to explore the performance effects of training such a model in various combinations of synthetic and real images.

We present our research on synthetic image training in the context of a real world anomaly detection system, including the results of testing on a large set of annotated proprietary production images.  We believe the methodology presented here can be readily applied to other systems, and make the case that synthetic images can replace the real images and still achieve a potentially useful level of performance.

The remainder of this paper is organized as follows.
Section~\ref{sec:background} describes concepts related to deep learning systems, synthetic data generation, and the specific models used in this paper.
Section~\ref{sec:methodology_gen} describes the technologies and processes used to generate synthetic images.
Section~\ref{sec:methodology_train} shows how synthetic data can augment real images to improve model accuracy.
Section~\ref{sec:methodology_transfer} compares those results to techniques using pretrained models and transfer learning.
Finally, Section~\ref{sec:conclusions} summarizes our key results and addresses further questions remaining for exploration.


\section{Background and Related Work}
\label{sec:background}

Synthetic data can be used to train deep learning models in a number of ways.  

First, in one extreme the model may be trained with only synthetic images, which can be useful in models where acquiring examples of desired detection conditions can be time consuming or unsafe.  For example, sufficient examples of rare flaws in products on an assembly line could be time consuming to capture for a quality control model, and examples of unsafe conditions may be challenging to acquire for a video surveillance system.  There has been some success with using purely synthetic data to train models \cite{richter_playing_2016,su_render_2015,hinterstoisser_annotation_2019}, and may be a good option depending on the use case.  Real images, if they exist, can be used as all or part of the test set to prove the model's accuracy.

Next, synthetic images may be mixed with real images in some combination, augmenting the size and/or variation of the training set presented to the model.  In published research, this method has been used to successfully decrease model training cost or improve model accuracy, and in some cases both\cite{peng_learning_2015,dwibedi_cut_2017,sun_virtual_2014}.

Finally, synthetic images can also be used to pretrain a model in a two-stage process, either by fitting a model to the synthetic set and then increasing the model bias toward real world examples by iterating over the real image set, or in a multi-model system such as DoubleU-net\cite{jha_doubleu-net_2020}, which is the primary focus of this paper.  This method is similar to using generalized image sets to pretrain a system (such as robotic vision) on patterns common to the real world, and then secondary training to adapt the model to a specific environment.  [sort out citations for examples here]


Jhang et al.\cite{jhang_training_2020} demonstrated training a Faster R-CNN\cite{ren_faster_2016} object detection model using synthetic images annotated with Unity Perception and generated at scale with Unity Simulation.  They found that while a model trained purely on a large (400,000) set of synthetic images performed poorly at detecting objects in situations with occlusions and low lighting, augmenting the synthetic images with a small number of real images significantly improved the detection accuracy over a model trained purely on a small (760) set of real images.  Their work was inspired by and complements findings from Hinterstoisser et al.\cite{hinterstoisser_annotation_2019}, who described a method for domain randomization by composing a backdrop of random objects in front of which the objects of interest are rendered and labeled.  Our process is distinguished by using a randomly oriented "skybox" surrounding the subject of interest, which achieves domain randomization with lowered scene complexity and randomized reflections.

Another method of domain adaptation to insert simulated objects of interest into real images, such as in \cite{yan2017sim}.

Rendered images of 3D scenes have been used to train object detection models for a long time, as exemplified by \cite{nevatia_description_1977} and \cite{lowe_three-dimensional_1987}.  More recently, advances in 3D rendering techniques have made photorealistic image generation practical.  \cite{hodan_photorealistic_2019} \cite{zhang_physically-based_2017} \cite{li_cgintrinsics_2018}

Other researchers have applied full domain randomization\cite{tobin_domain_2017} to synthetic image generation with varying degrees of success.  \cite{hinterstoisser_annotation_2019,tremblay_deep_2018,borrego_applying_2018}.
Our approach is a hybrid between full domain randomization and photorealistic rendering, varying the lighting and subject/background orientation and random sampling from a set of realistic textures.  The approaches described in \cite{mitash_self-supervised_2017}, \cite{prakash_structured_2019} and \cite{tremblay_training_2018} are most similar to our own in this regard.


Successful specialization of U-net models has been achieved \cite{iglovikov2018ternausnet, frid2018improving} using VGG encoders\cite{simonyan_very_2014} pretrained on the ImageNet\cite{deng_imagenet_2009} dataset.

Recent research has shown that models developed using synthetic data can be used as a basis for more specific models.  This \textit{transfer learning} can be used in many tasks, such as enhancing detection of object position in \cite{inoue_transfer_2018, yan2017sim} and separating target objects from visual distractors in \cite{zhang2017sim}.






\section{Synthetic Image Generation}
\label{sec:methodology_gen}
In this section we describe the tools and workflow developed for creating synthetic images, followed by our experiment designs for validating the output images and using the generated data to train a deep learning model.  Software used includes Unity 2020.1, Unity High Definition Rendering Pipeline (HDRP) 7.4.1, and PiXYZ Plugin 2019.2.1.14.

\subsection{3D Modeling}

The image generator was built as a set of scene descriptions, models, and scripts in the Unity 3D game development platform. 

For our use case, a vehicle model was translated from its native CATIAv5 CAD format [check name] into a Unity asset with the PiXYZ plugin.  Importing the CAD object was relatively labor intensive due to a technical difficulty in mapping part materials to Unity textures, which is an area of current work.  The work-around for our purposes was to manually assign textures to the approximately 10,000 visible surfaces in the imported Unity asset.


\subsection{Realistic Rendering}

In general, synthetic images for model training need to exemplify the characteristics of real images that the model relies on for accurate classification.  While these qualities could be vastly different depending on the model, for our use case we needed images that embody the broad range of shadows and reflections seen in the production environment.  Rather than attempting to identify and optimize for the most important image features, our approach was to create images as accurately as possible with a goal of being indistinguishable from real images by a human observer. 

Images were rendered using the Unity High Definition Rendering Pipeline.  We relied on a number of Unity features designed for high rendering accuracy, and avoided many approximation features designed to improve rendering performance in a game setting requiring high framerate with limited hardware resources.  Our settings for image accuracy were largely influenced by guidelines from Unity\cite{hdrp_visuals}.  The Unity "camera" object was configured to mimic the properties of the physical camera used to capture real images.  A full disclosure and justification of the rendering settings we used would be lengthy and beyond the scope of this paper, and will be made available on publication.

We found that several external resources were very helpful in creating realistic image rendering, especially in our use case with automotive models.
In particular, Unity's automotive industry-focused Measured Materials library\cite{measured_materials} helped us simulate the paint, glass, rubber, and plastic textures of a real vehicle.
Skyboxes were sampled from the Unity HDRI pack, captured using techniques described by Lagarde et al.\cite{lagarde2016hdri}.

\subsection{Domain Randomization}

We chose a hybrid approach to domain randomization, rendering the image subject as accurately as possible with ambient lighting similar to the production environment.  Randomized attributes included subject position relative to the camera within plausible constraints, vehicle exterior paint colors from a set of possible values, and a single light source (the sun) with varying position.

To separate the subject from the background, we used a background skybox with a very "busy" texture, and then randomized its orientation on all 3 axis for every scene.  This served a secondary purpose in creating randomized reflection patterns on all surfaces of the vehicle.

Randomization of objects in the scene was accomplished with a set of scripts written in C-sharp, used natively in Unity for game logic.  



\subsection{Segment Labeling}

We labeled image segments by capturing multiple images from each randomized scene -- one fully rendered image, and then one false color image for each segment.  This could have been achieved in many ways, but the approach we found to be most performant in Unity was to maintain a second "mask" copy of the subject model completely colored with an "unlit" black texture, locked to the same position as the color model.  Two identical cameras in the same position were used, one able to see the color model, background, and lighting and the other camera only able to see the mask model.  

\begin{figure*}[ht]
    \centering
    \includegraphics[width=.48\textwidth]{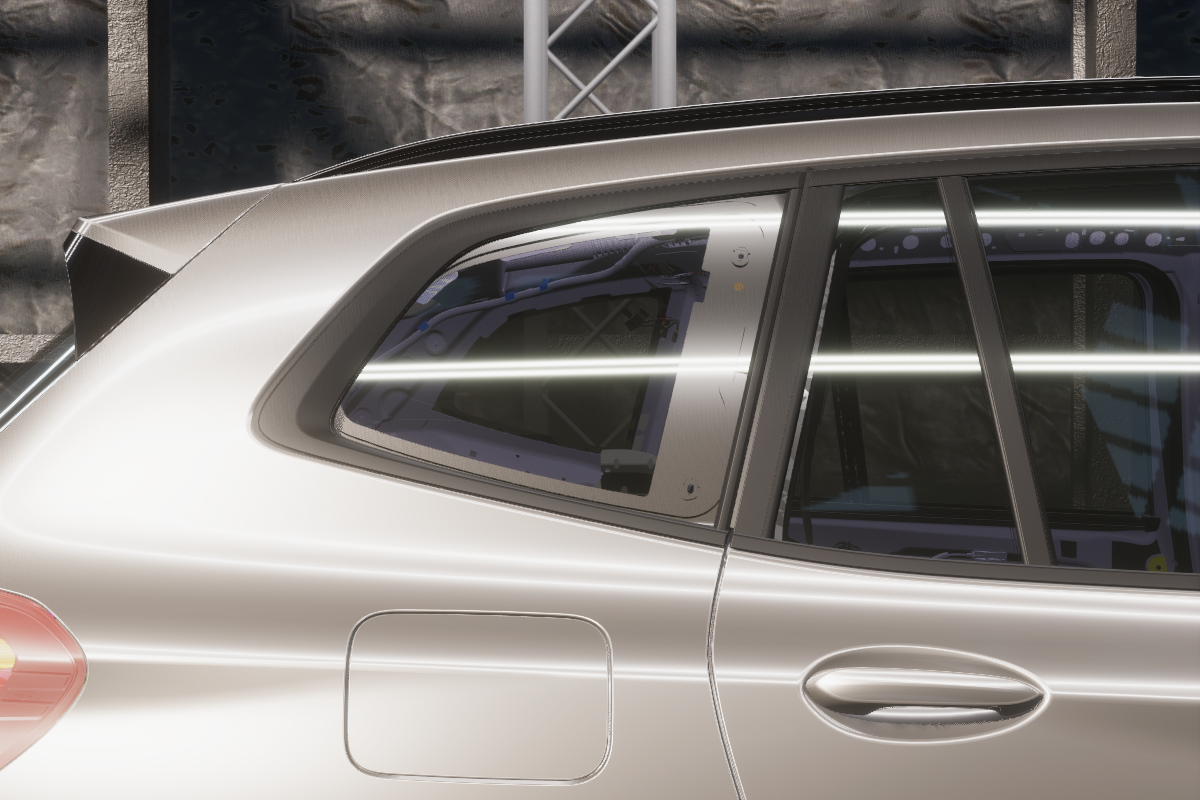}\hfill
    \includegraphics[width=.48\textwidth]{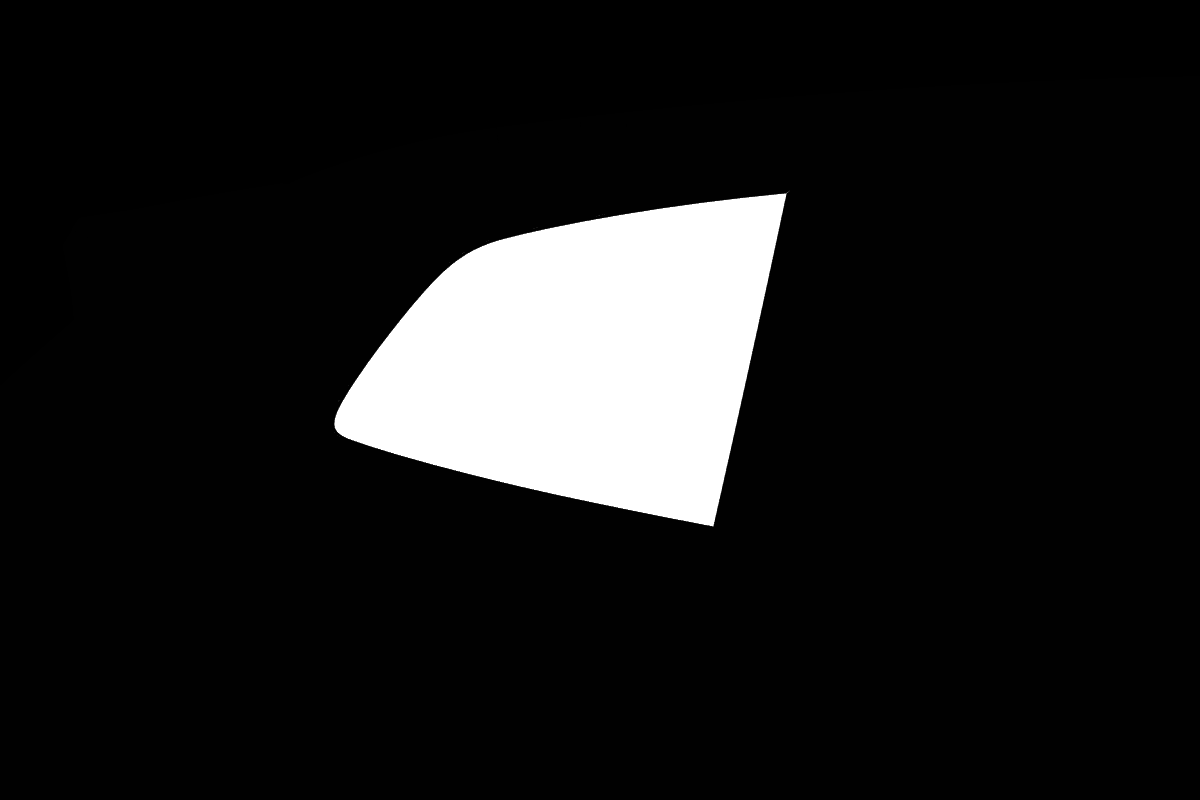}
    \\[\smallskipamount]
    \includegraphics[width=.48\textwidth]{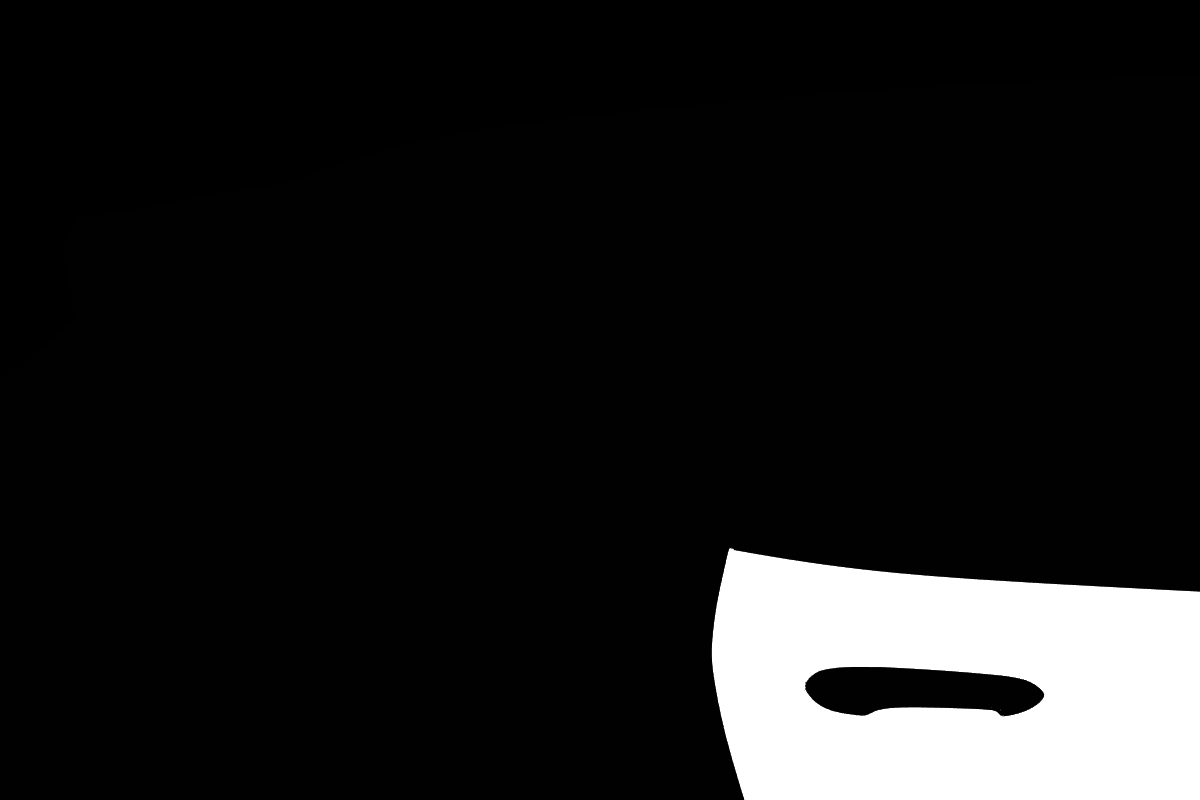}\hfill
    \includegraphics[width=.48\textwidth]{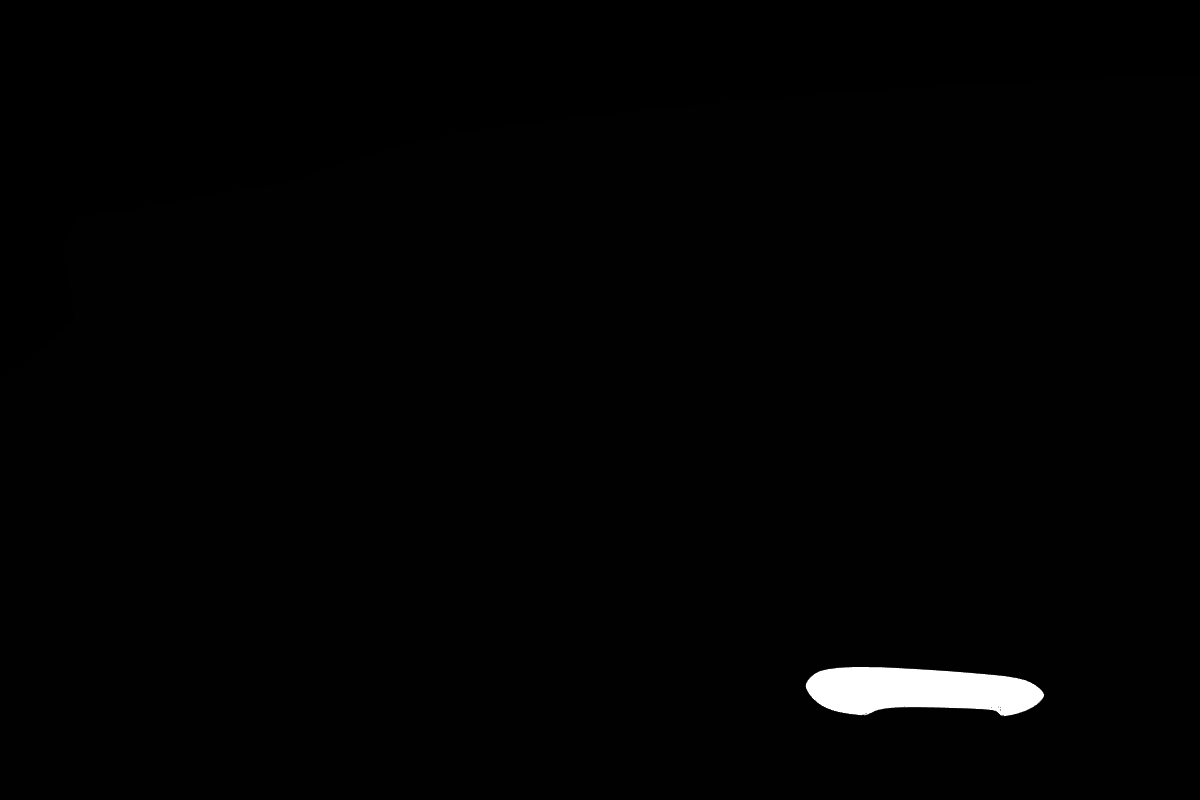}
    \caption{A 3D generated image (top left) in addition to a series of one-hot encoded masks segmenting each object class.}
    \label{fig:segment_labeling}
\end{figure*}

After the normal image was captured with the color camera, the segment capture phase would iterate through groups of components comprising each segment, recolor the group with an unlit white texture, capture an image with the mask camera, and then recolor the group to the unlit black texture.  Figure \ref{fig:segment_labeling} shows the resulting image segments.  This approach had the performance advantage of minimizing the retexturing of materials on the model.  This also allowed us to capture occlusions by components not part of the segment of interest, such as the door handles in the example images.

\section{Model Training}
\label{sec:methodology_train}

To validate the effectiveness of the synthetic image generator, we conducted experiments comparing models trained with varying amounts of real labelled images augmented with synthetic data.  
Our available data consisted of 14,125 labelled images of real vehicles in a production line, each of which contained one or more examples of eight distinct feature classes.
From this dataset, a 10\% holdout set was randomly selected for validating models, leaving 12,712 images in the real dataset $R$ for training.
The frequency of each feature's appearance is described in Table~\ref{tab:dataset_frequency}, where the subset of the real image set $R$ with one or more pixels belonging to a feature class $f$ is given as $R_f = \{e|e \in R$~and~$f\in e\}$, and an example frequency of $|R_f|/|R|$.

Using the synthetic image generator described in Section~\ref{sec:methodology_gen}, we rendered a set of 40,406 synthetic images and labels $S$ with the same feature classes as $R$.  Due to a slightly smaller horizontal range of camera freedom, some classes were represented more or less heavily in the synthetic set, as detailed in Table~\ref{tab:dataset_frequency}.  However, as we weight each class equally in our metrics and present aggregate statistics over the entire dataset, we deemed that the example frequency weights would not affect the conclusions.

\begin{table}[t]
\centering
\caption{Feature Example Frequency in Image Sets}
\label{tab:dataset_frequency}
\begin{tabular}{c c c c c}
    \hline
    & \multicolumn{2}{c}{real images $R$} & \multicolumn{2}{c}{synthetic images $S$} \\
    \hline
    feature & examples & frequency & examples & frequency \\
    \hline
    back door    & 5,994 & 47.09\% & 40,231 & 99.57\% \\
    back window  & 5,854 & 45.99\% & 40,263 & 99.65\% \\
    rear window  & 4,844 & 38.05\% & 24,080 & 59.60\% \\
    front door   & 6,599 & 51.84\% & 22,308 & 55.21\% \\
    front window & 5,985 & 47.02\% & 26,171 & 64.77\% \\
    door handle  & 4,670 & 36.69\% & 40,084 & 99.20\% \\
    mirror       & 3,897 & 30.62\% & 6,932 & 17.16\% \\
    tail light   & 4,511 & 35.44\% & 8,501 & 21.04\% \\
    \hline
\end{tabular}
\end{table}

\subsection{Training Methodology}

Images and labels were used to train U-net\cite{ronneberger_u-net_2015} convolutional neural network models implemented in TensorFlow\cite{tensorflow2015-whitepaper} 2.0.0 and Keras\cite{chollet2015keras} 2.2.4-tf.  Models were trained using an NVIDIA DGX-2 with Tesla V100 GPUs running Ubuntu 18.04.4 LTS.  

In this section, all U-net model structure and parameters are identical with the exception of input datasets.  The U-net implementation was derived from code provided by Debesh et al. in their Double-U-net supplement, to be consistent with the further work in Section~\ref{sec:methodology_transfer}.  From the original U-net description, the only significant difference is the use of batch normalization\cite{ioffe2015batch} after the convolutional layers along the contracting path, which resulted in more consistent training and better generalization in our use case.

A hyperparameter search using real and synthetic datasets revealed optimal parameters that were similar enough to avoid differentiation between the domains.  As the purpose of this work is to explore the tradeoffs of synthetic vs real data, we chose parameters that resulted in consistent and stable training sessions rather than strictly optimizing for the highest possible accuracy.  For our datasets, a dropout probability of 0.30, a batch size of 64, and a learning rate of 0.0020 resulted in models that converged quickly and consistently within a reasonable limit on training time and generalized well to the validation data.

As synthetic data can be seen as a form of data augmentation, we chose to forego any traditional augmentation techniques (randomized cropping, gamma shifts, etc.) to present clear results, with the single exception of randomly flipping all training images horizontally to match the real dataset's imaging of both sides of the vehicle.
During training, models were evaluated each epoch against the disjoint validation set.  To prevent overfitting, we used an early stopping mechanism to halt training and revert to the best weights if no improvement in validation set prediction loss was made over 30 epochs.

\subsubsection{Metrics}

While the image generation and training techniques share applicability with object detection and instance segmentation models with more actionable metrics, we quantify the performance of a standard multiclass U-net segmentation model simply with per-pixel mean intersection-over-union (mean IoU) with uniform class weighting and a prediction threshold of 50\%.


\subsection{Real Dataset Supplementation}



\newcommand{\w}{0.94}

\begin{figure}[t]
    \sidesubfloat[]{
        \includegraphics[width=\w\textwidth]{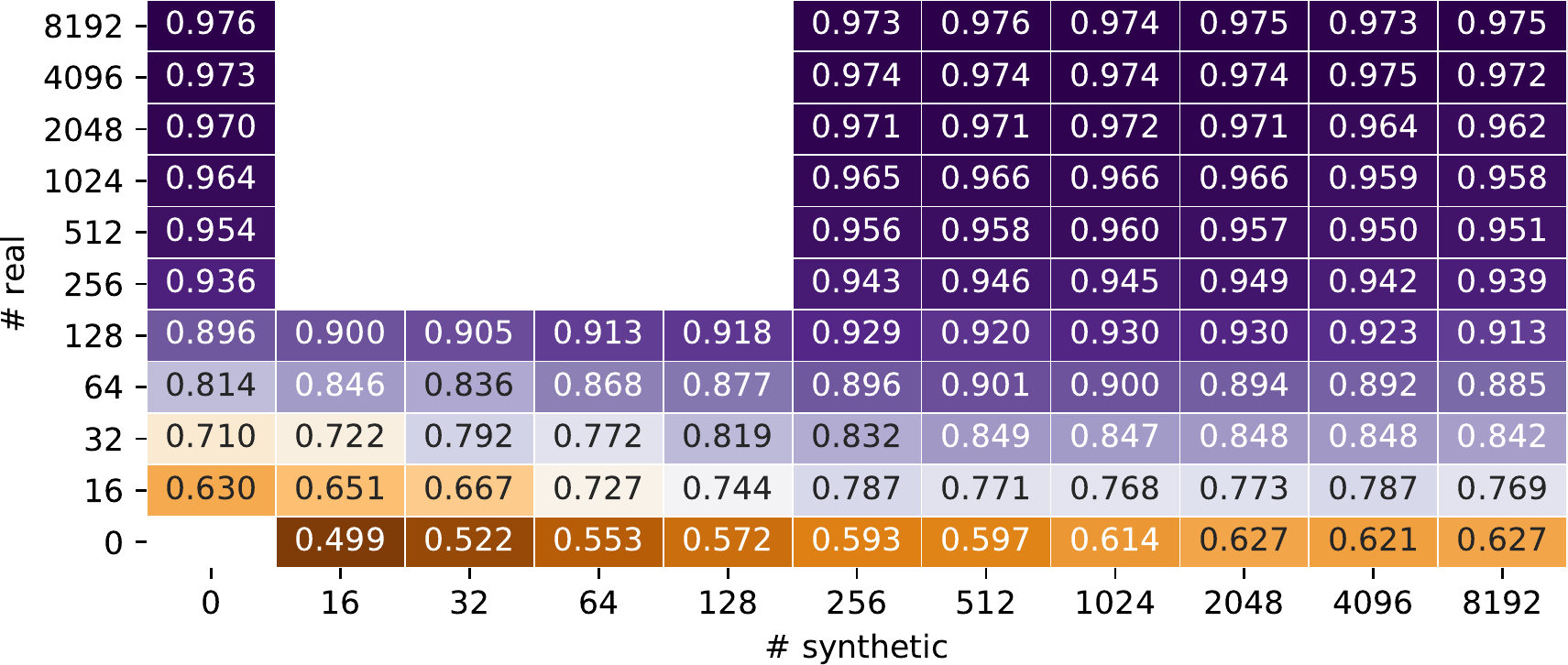}
        \label{fig:real_synth_matrix_iou}
    }
    \\
    \vspace{0.5cm}

    \sidesubfloat[]{
        \includegraphics[width=\w\textwidth]{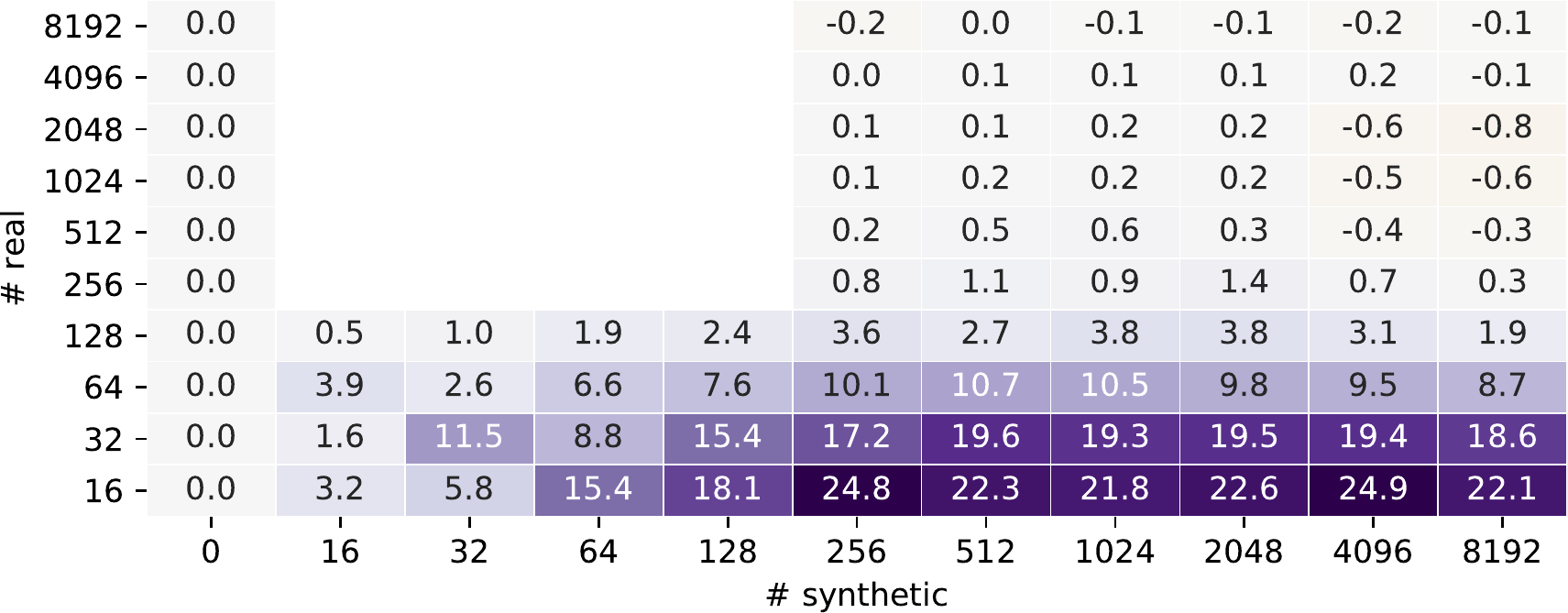}
        \label{fig:real_synth_matrix_increase}
    }
    \\
    \vspace{0.5cm}

    \sidesubfloat[]{
        \includegraphics[width=\w\textwidth]{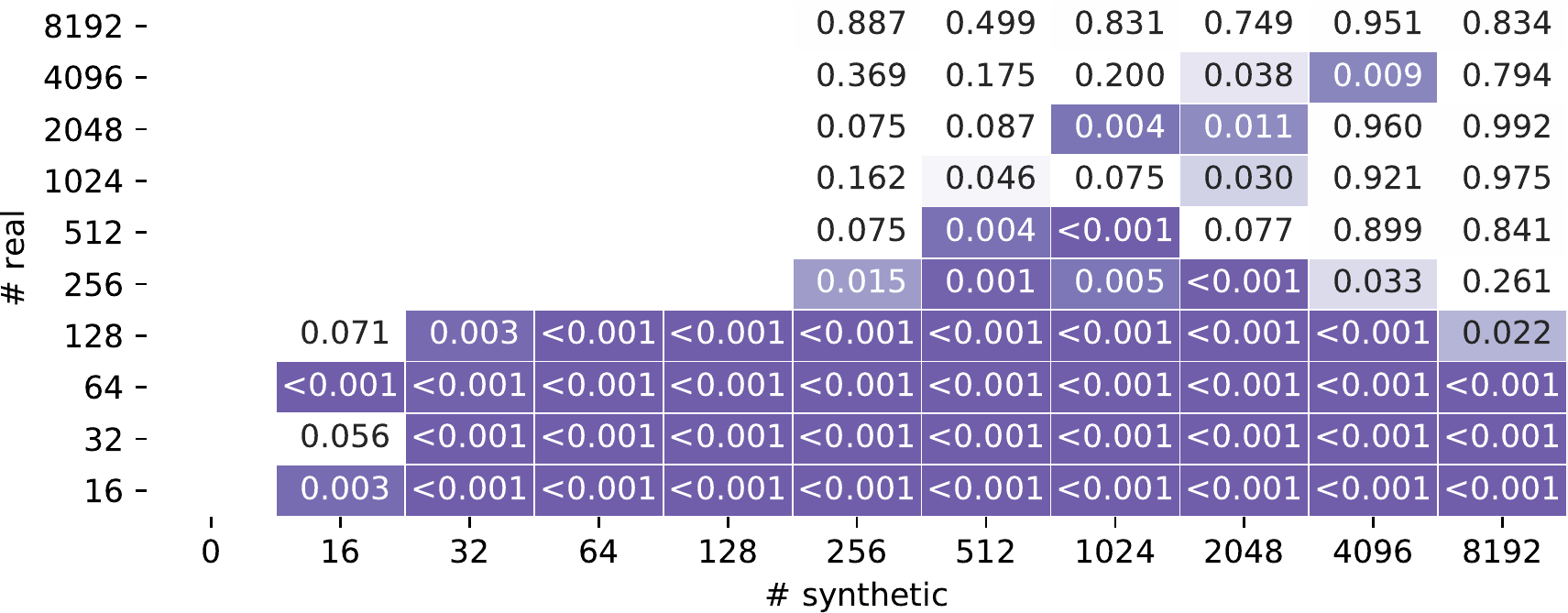}
        \label{fig:real_synth_matrix_pvalue}
    }

    \caption{
        Aggregated mean prediction IoU \protect\subref{fig:real_synth_matrix_iou} of U-net models trained on random samples from real and synthetic datasets.
        Models augmented with synthetic data showed up to 24.9\% higher prediction accuracy \protect\subref{fig:real_synth_matrix_increase} than the baseline, particularly with limited amounts of real training images.
        The $p$-values \protect\subref{fig:real_synth_matrix_pvalue} of one-sided T-tests, $H_a : \overline{IoU}(M_{r,s}) > \overline{IoU}(M_{r,0})$, show significant accuracy increases ($p\leq 0.05$, highlighted) in most models trained with 256 or fewer real images.
    }
    \label{fig:real_synth_matrix}%
\end{figure}

To determine how supplementing a dataset of real images with synthetic images would affect model training and accuracy, we trained instances of multiple model classes with different mixtures of images from both sets.
Subsets of the real image set $R$ of sizes $N=\{0,16,32,...,8192\}$ were paired with subsets of the synthetic image set $S$ from the same size range, forming the axes of the 11x11 matrices shown in Figure~\ref{fig:real_synth_matrix} with model classes at each intersection.
For each model class, random samples from $R$ and $S$ were used to train individual U-net segmentation models with parameters reported above.
The number of models trained in each class was sufficient that the confidence interval ($\alpha =0.95$) width of the mean truth/prediction IoU measurements on the real image validation set was less than 5\% of the mean value, requiring between 7 and 30 model instances for each image set size pair.
We refer to the resulting set of segmentation models as $M$, where $m_{r,s,i}\in M: r\in N, s\in N, i\in[0..|M_{r,s}|)$ is one instance of a class of U-net models trained on $(r,s)$ random images from datasets $R$ and $S$, and we report aggregate statistics over the model class $M_{r,s}$ at each cell in the matrices of Figure~\ref{fig:real_synth_matrix}.


Figure~\ref{fig:real_synth_matrix_iou} aggregates the mean IoU predictions of each trained model class on the unseen validation set from the real image domain.  We observe a general trend of increasing accuracy with larger samples of real images, with diminishing returns as the training images grow to sufficiently represent the domain features.  Along the horizontal axis, we see that augmentation with synthetic data tended to increase accuracy, with greater yields in models trained on smaller real datasets.  We also observe that models trained on purely synthetic data tend to poorly predict the real domain, even with thousands of examples.


\begin{figure}[t]
    \centering
    \includegraphics[width=\textwidth]{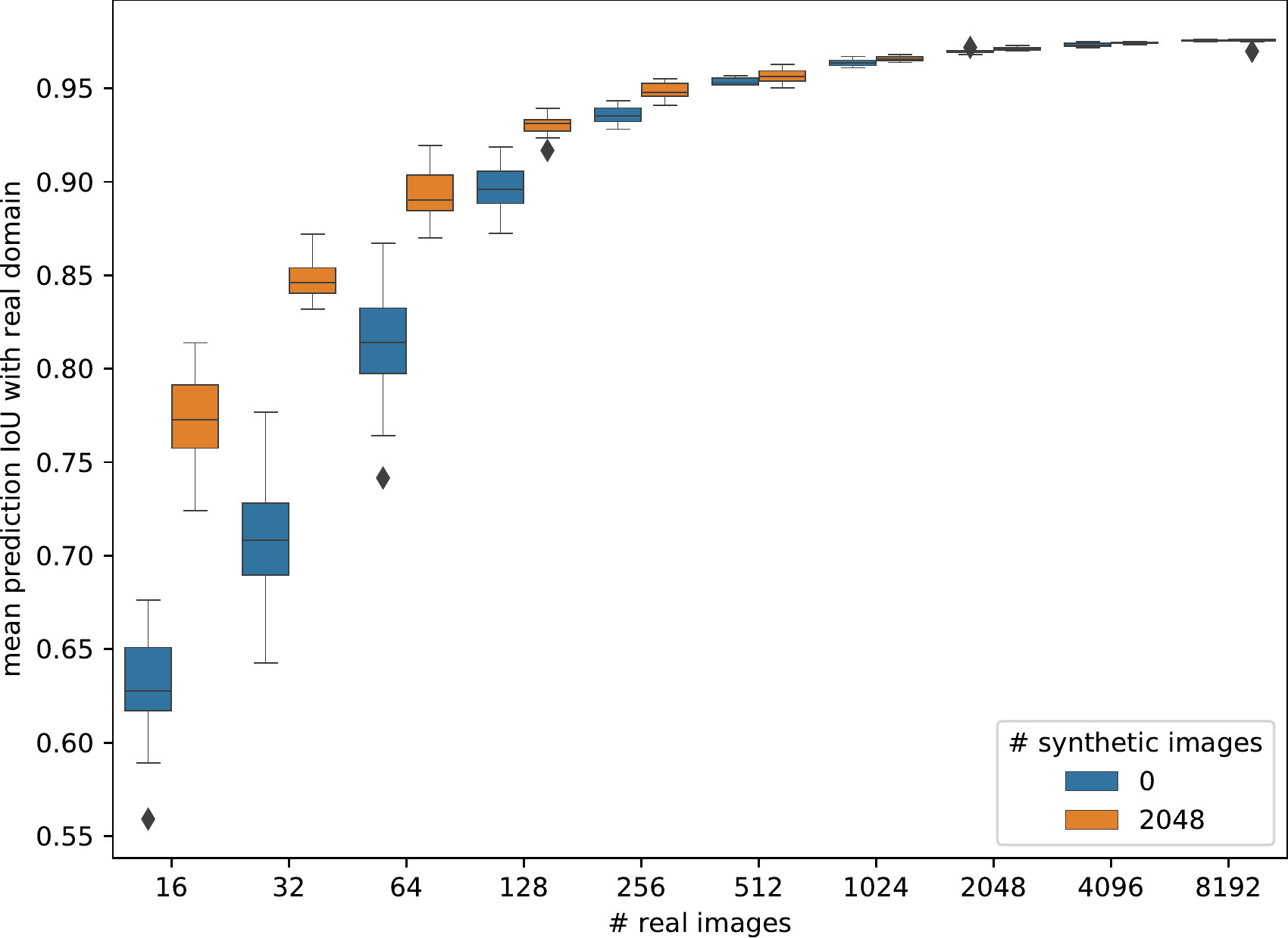}\hfill
    \caption{Mean IoU of model predictions on a validation set of 1176 real images.  Each IQR plot describes between 7 and 30 individual U-net models trained on random subsets of real and synthetic images.  In general, augmenting smaller ($\leq 256$) sets of real images resulted in higher accuracy and less variation in the trained models, with diminishing returns as the real data became sufficiently representative of the domain.}
    \label{fig:real_synth_variance}
\end{figure}

To discuss the results of synthetic data augmentation, we first look at the effects of augmentation on model reliability.
Figure~\ref{fig:real_synth_variance} shows the summary statistics of mean validation set predictions for model classes trained on purely real images and those augmented with 2048 synthetic images, which details columns 0 and 2048 from Figure~\ref{fig:real_synth_matrix_iou}.
Models trained with smaller random samples of real images tended to show more variation in their resulting prediction accuracy.
We observe that augmentation tended to increase mean accuracy and decrease variance in models trained with less than 256-512 real images.


Augmenting the real training sample with varying amounts of synthetic data yields better results, depending on how accurate the model is to begin with.
Figure~\ref{fig:real_synth_matrix_increase} reshapes the data in Figure~\ref{fig:real_synth_matrix_iou} as a percentage increase in mean prediction IoU relative to that of the pure real set (column 0).
We can see that augmenting models trained with 512 or more real images only results in a marginal increase, at best 0.6\%.
However, in models trained with 256 or fewer real images, the accuracy increase is substantial, up to 25.0\% when only 16 real images are available.
We can also see that the addition of any amount of real images results in models that are more accurate than those trained on synthetic data alone.
This is supported by the $p$-values of one-sided T-tests, $H_a : \overline{IoU}(M_{r,s}) > \overline{IoU}(M_{r,0})~\forall~r,s\in N$, shown in Figure~\ref{fig:real_synth_matrix_pvalue} with $p < 0.05$ highlighted.


\begin{figure}[t]
    \centering
    \includegraphics[width=\textwidth]{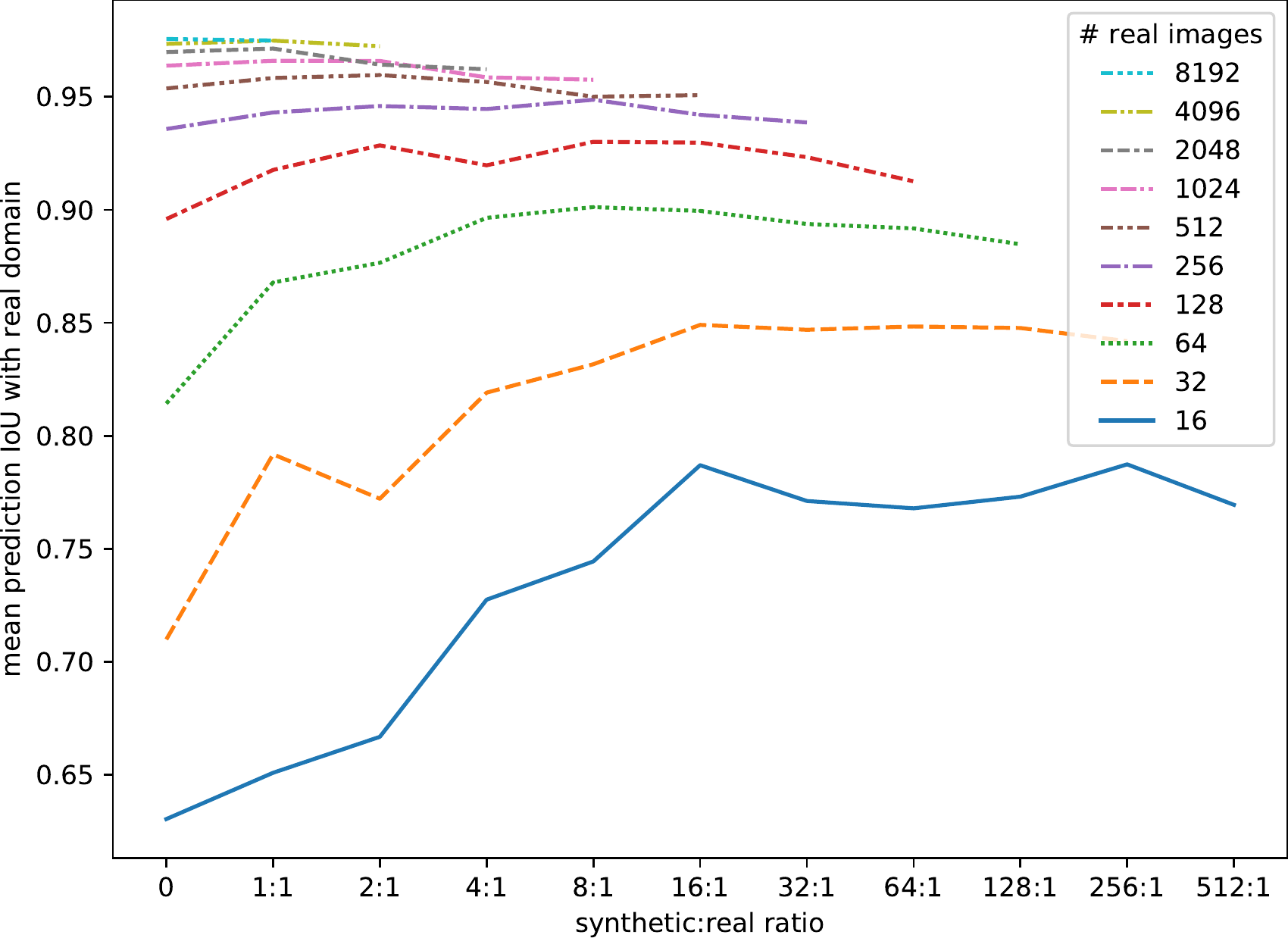}\hfill
    \caption{Mean prediction IoU of U-net models on real images, viewed by the ratio of real to synthetic data in the training datasets.  Each trend exhibits an inflection point where accuracy decreased, presumably due to limited capacity of the model to encompass both the real and synthetic domain.}
    \label{fig:real_synth_ratio}
\end{figure}

Figure~\ref{fig:real_synth_matrix_increase} also shows that in some cases, particularly in those with 512 or more real images, the addition of large amounts of synthetic data correlate with a slight decrease in prediction accuracy, presumably due to dilution of the samples from the real domain and a limited capacity of the model to encompass both the real and synthetic domains.
We can observe this trend more clearly when viewing the relationship between real and synthetic image set sizes as a ratio, shown in Figure~\ref{fig:real_synth_ratio}.
Each real image set size exhibits an inflection point where accuracy declines, which we suspect is dependent on the capacity of the model and similarity between real and synthetic data in a particular use case.


To visualize the differences in prediction accuracy, Figure~\ref{fig:stacked_predictions} presents the segmentation maps predicted by 10 different models, trained on 16-256 real images and augmented with either 0 or 2048 synthetic images.
In contrast to the randomly selected images used to train the models Figure~\ref{fig:real_synth_matrix}, each real dataset larger than 16 images is a superset of the smaller datasets, and the same real datasets and 2048-image synthetic dataset are reused in each of the augmented models.
For this example image, the quality of the predictions are fairly low in the pure real models, limiting usefulness depending on the use case.
The addition of synthetic images results in clearly defined door/window boundaries with even the smallest real training set, and better identification of smaller features such as the door handles at 64 real images compared to requiring 128 without augmentation.

\renewcommand{\w}{0.158}

\begin{figure*}[t]
    \centering
    \frame{\includegraphics[width=\w\textwidth]{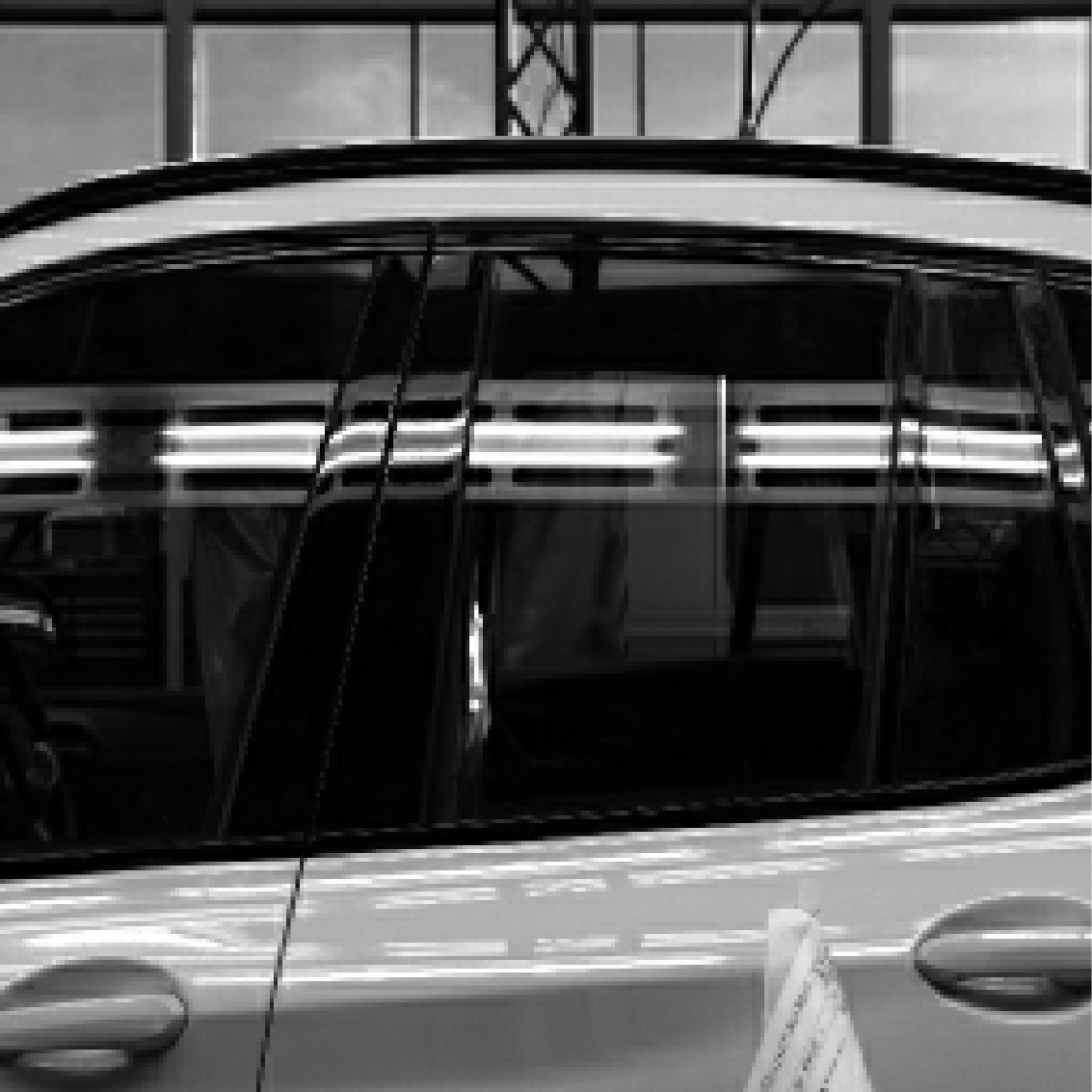}}
\hspace{1pt}
    \frame{\includegraphics[width=\w\textwidth]{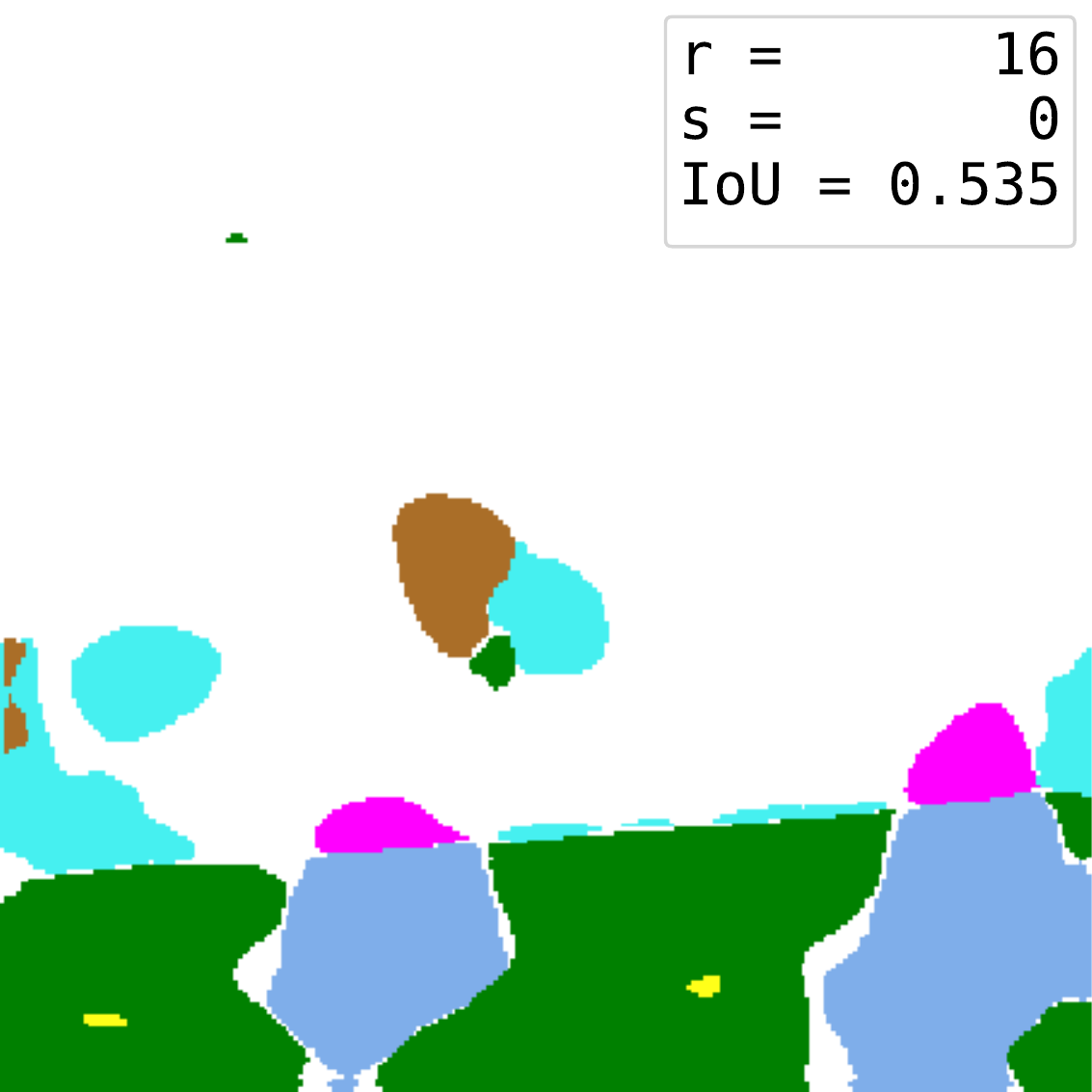}}
    \frame{\includegraphics[width=\w\textwidth]{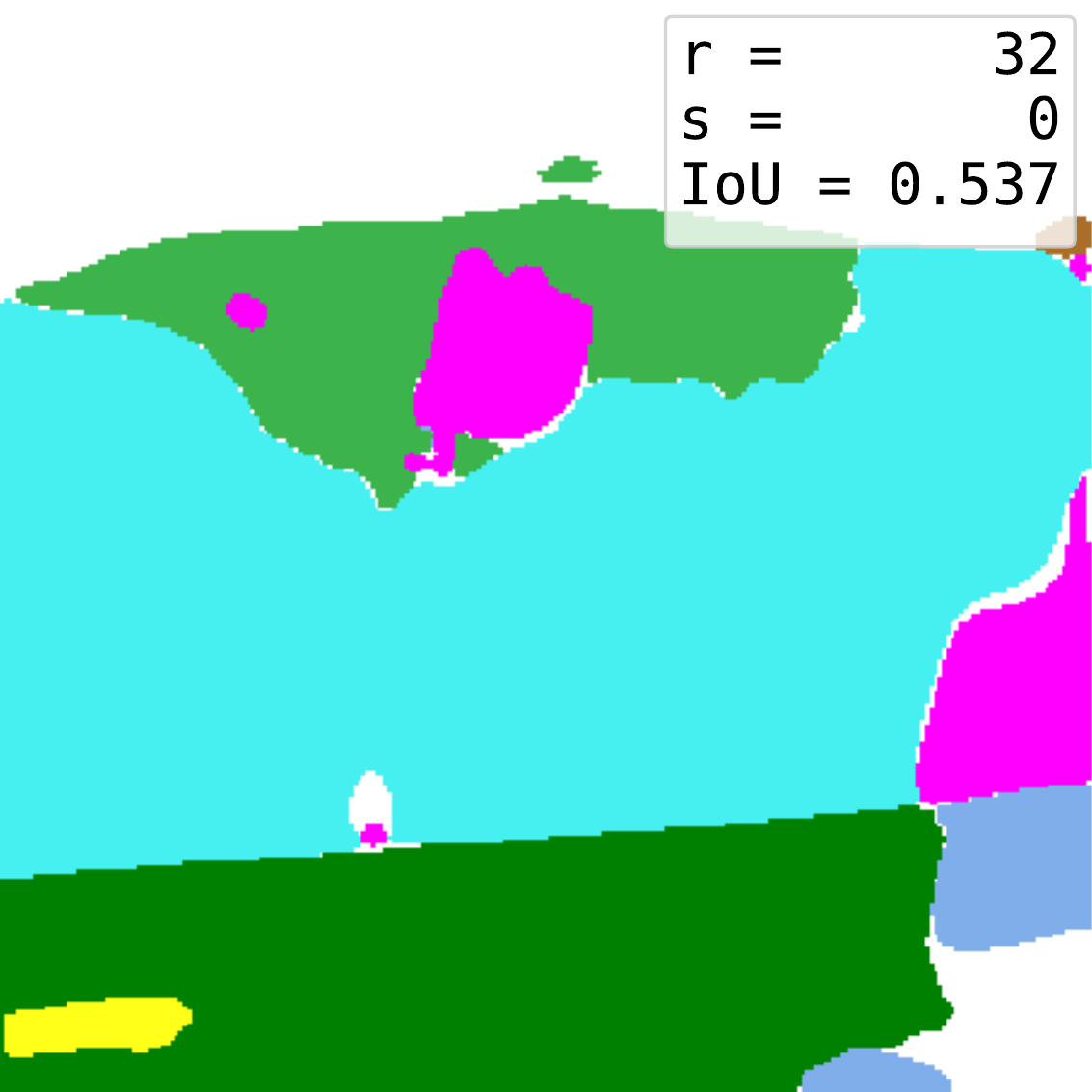}}
    \frame{\includegraphics[width=\w\textwidth]{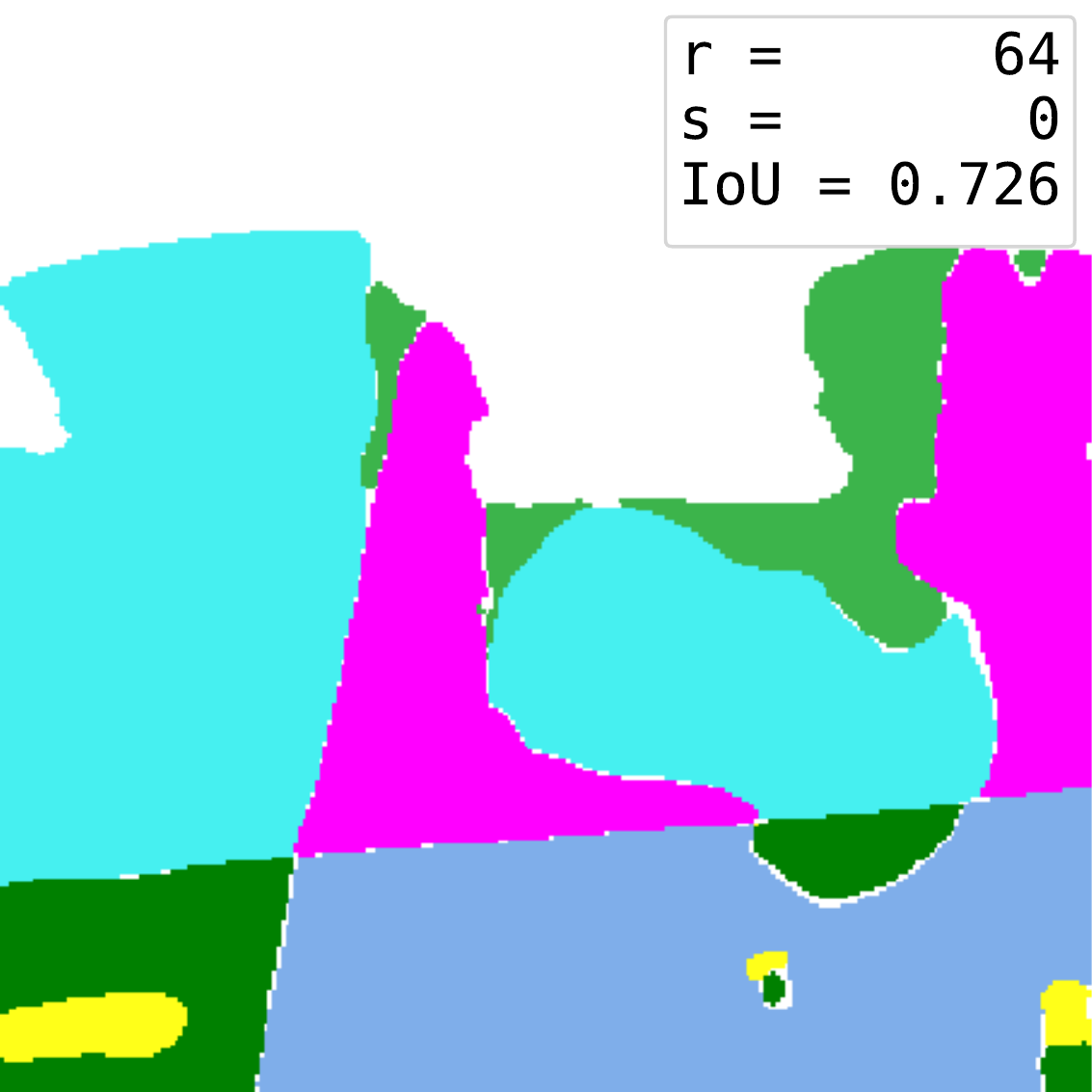}}
    \frame{\includegraphics[width=\w\textwidth]{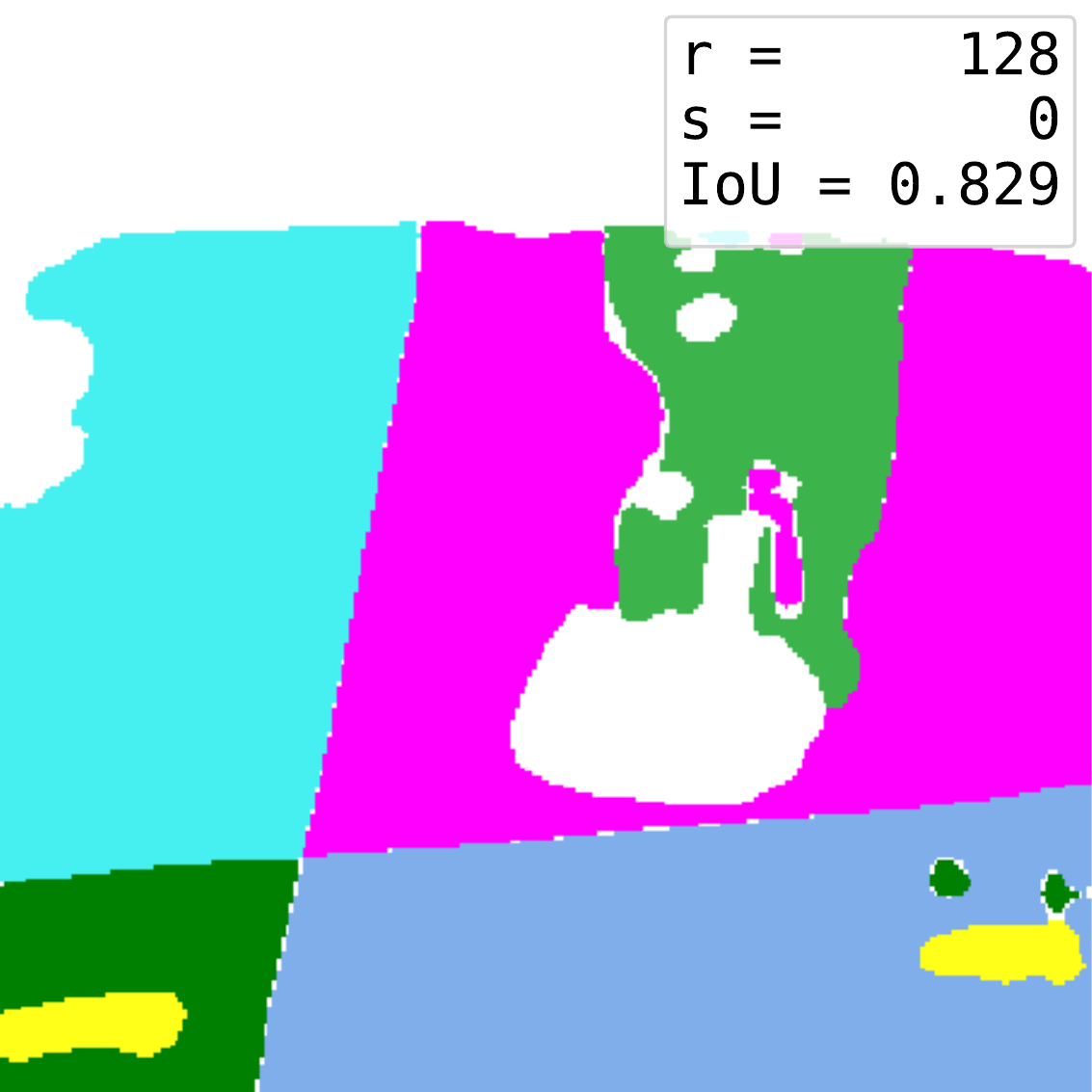}}
    \frame{\includegraphics[width=\w\textwidth]{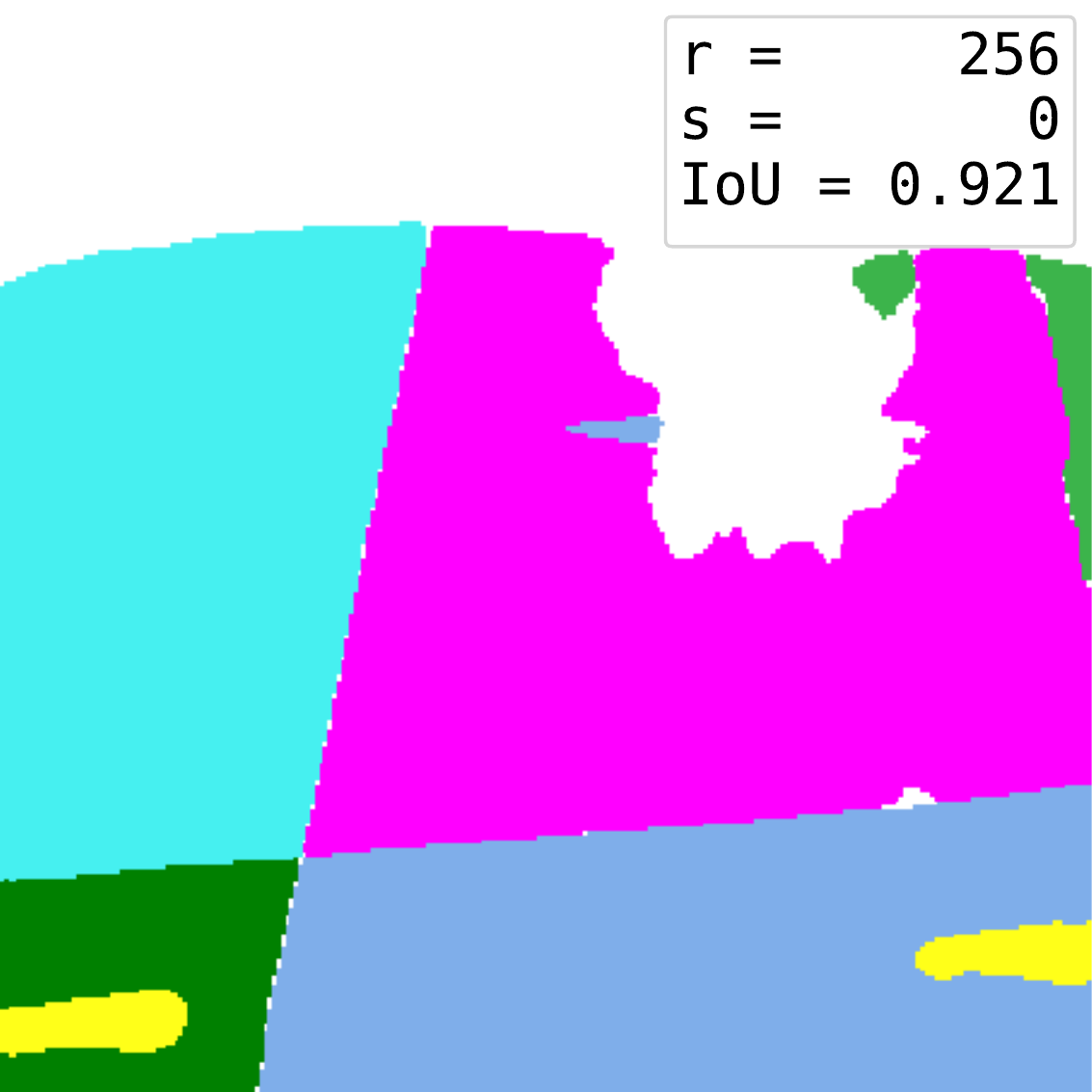}}
    \label{fig:stacked_predictions_pure_real}
\vspace{0.2pt}
    \centering
    \frame{\includegraphics[width=\w\textwidth]{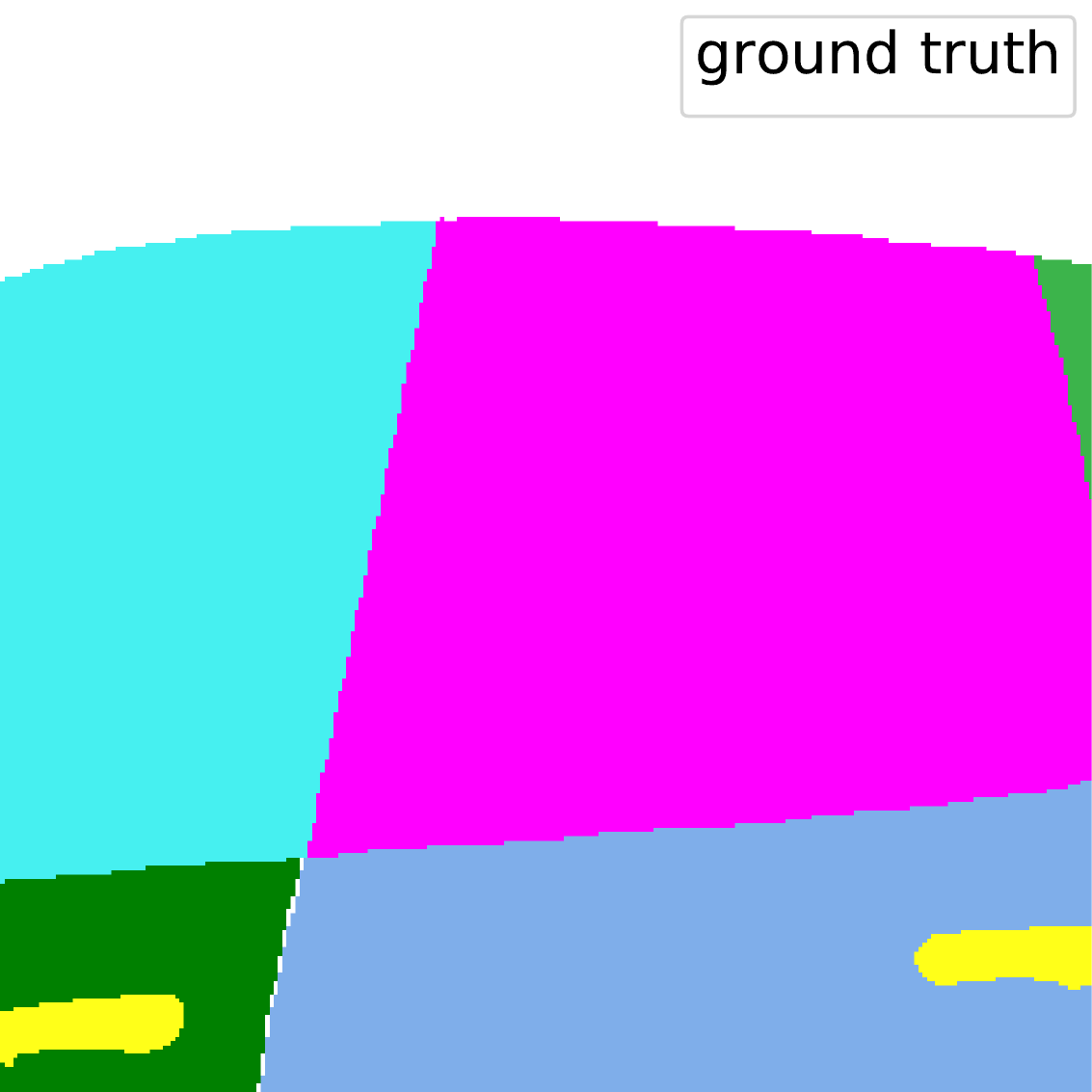}}
\hspace{0.5pt}
    \frame{\includegraphics[width=\w\textwidth]{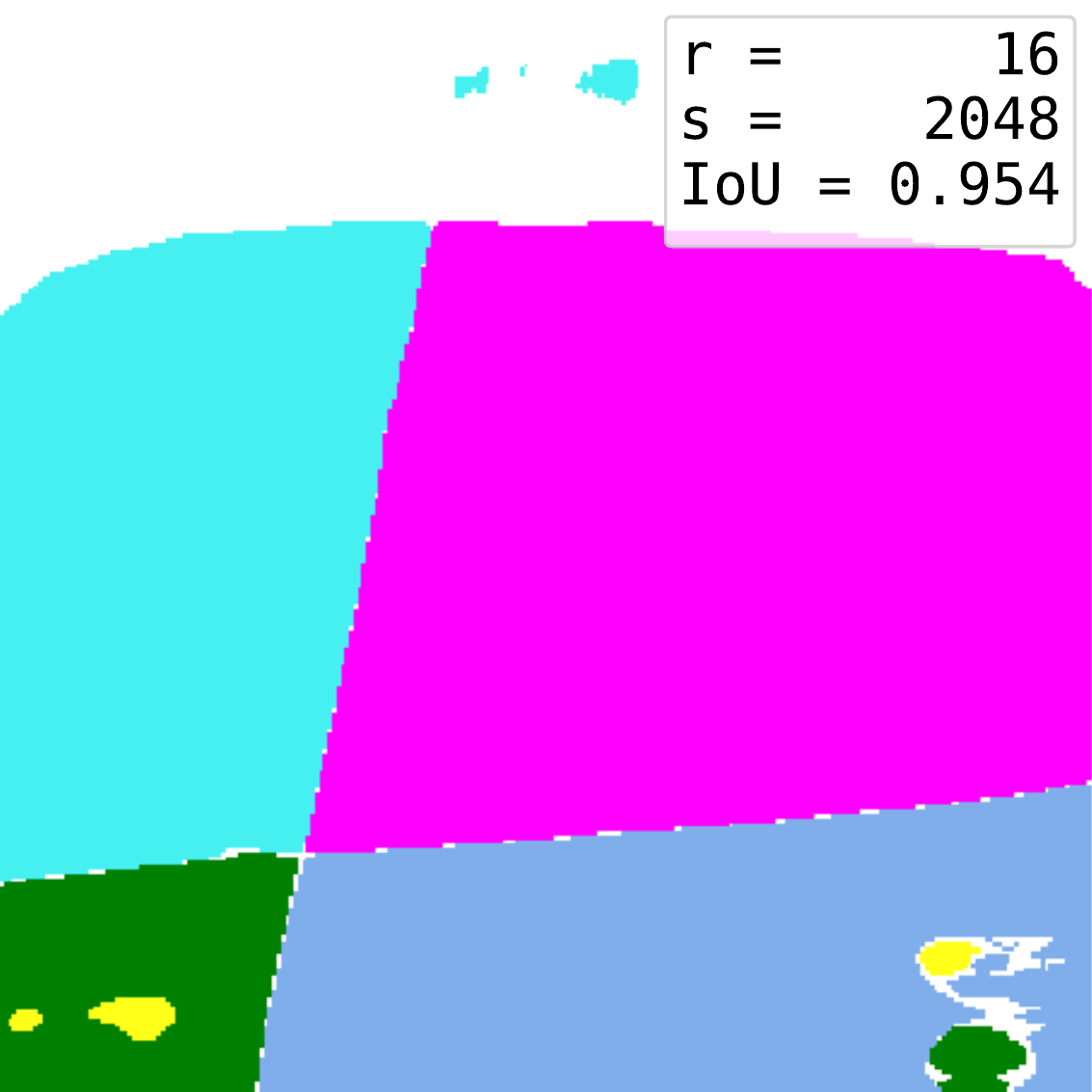}}
    \frame{\includegraphics[width=\w\textwidth]{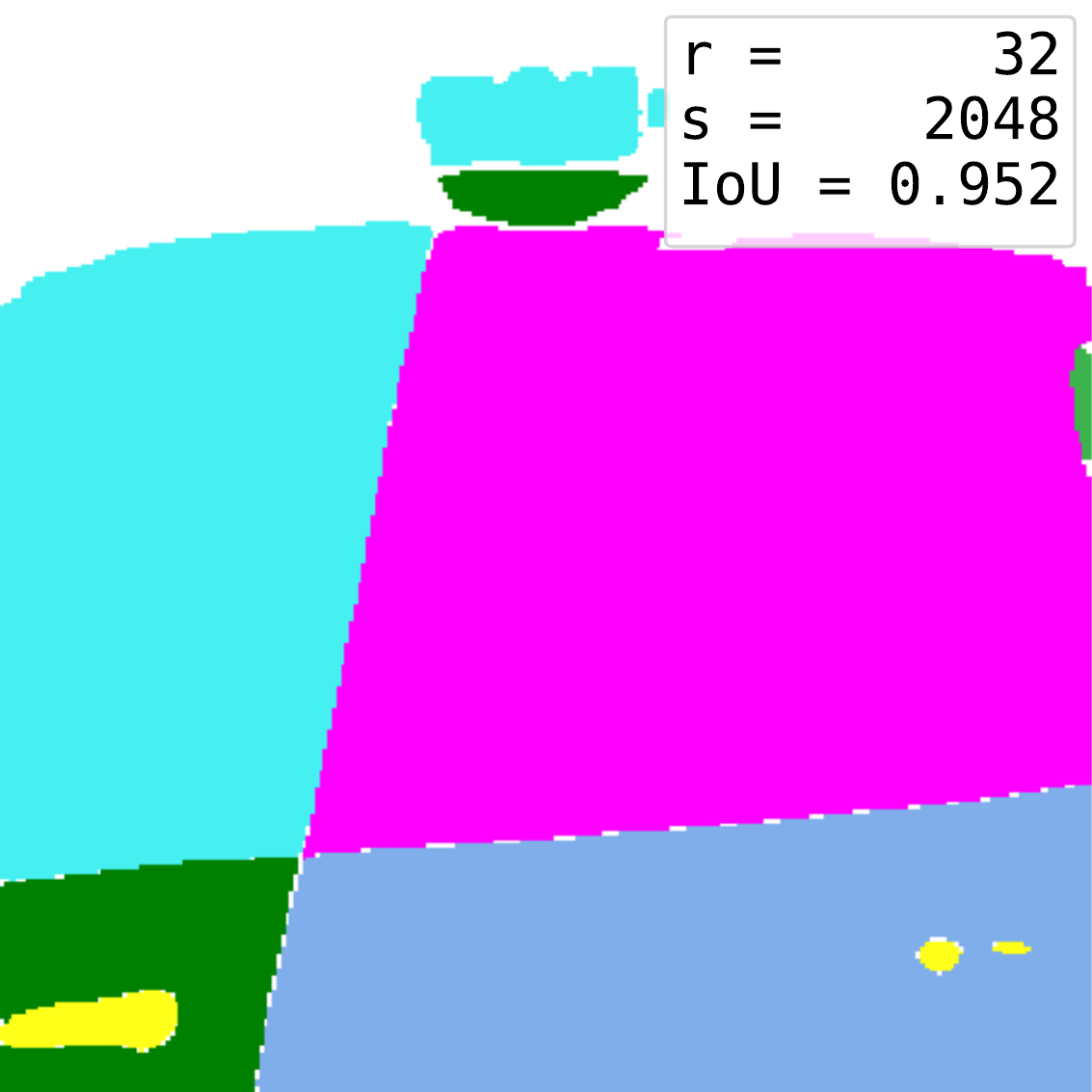}}
    \frame{\includegraphics[width=\w\textwidth]{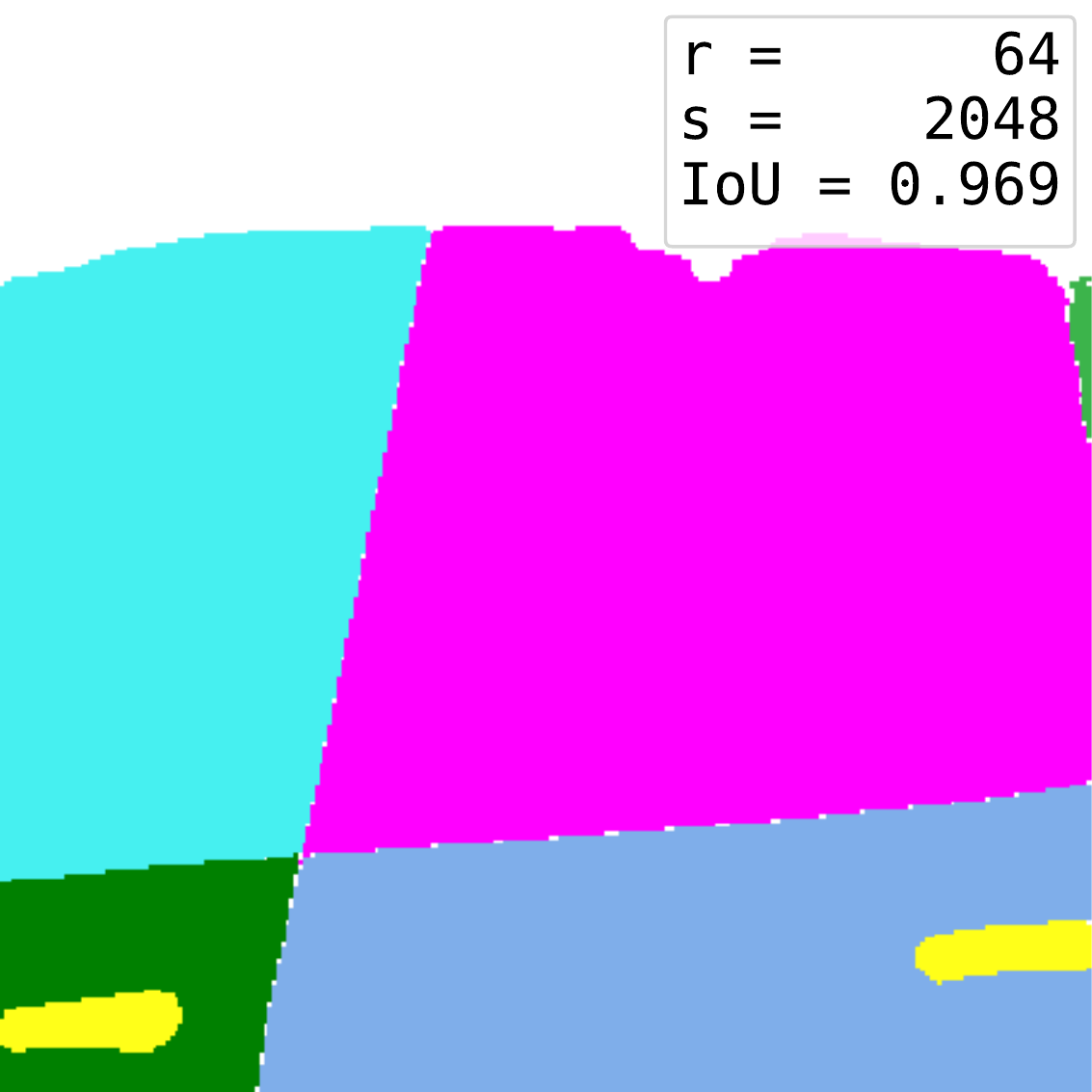}}
    \frame{\includegraphics[width=\w\textwidth]{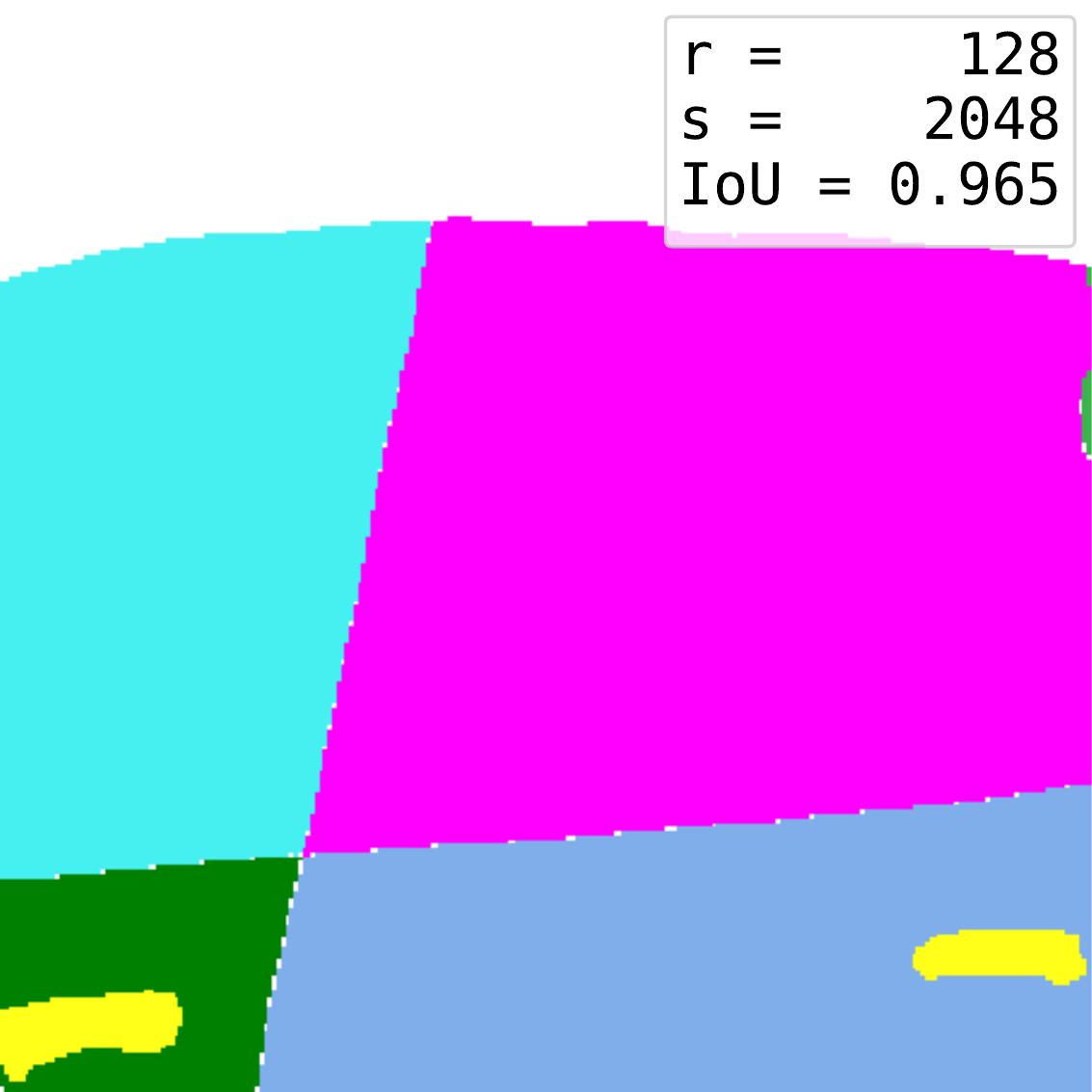}}
    \frame{\includegraphics[width=\w\textwidth]{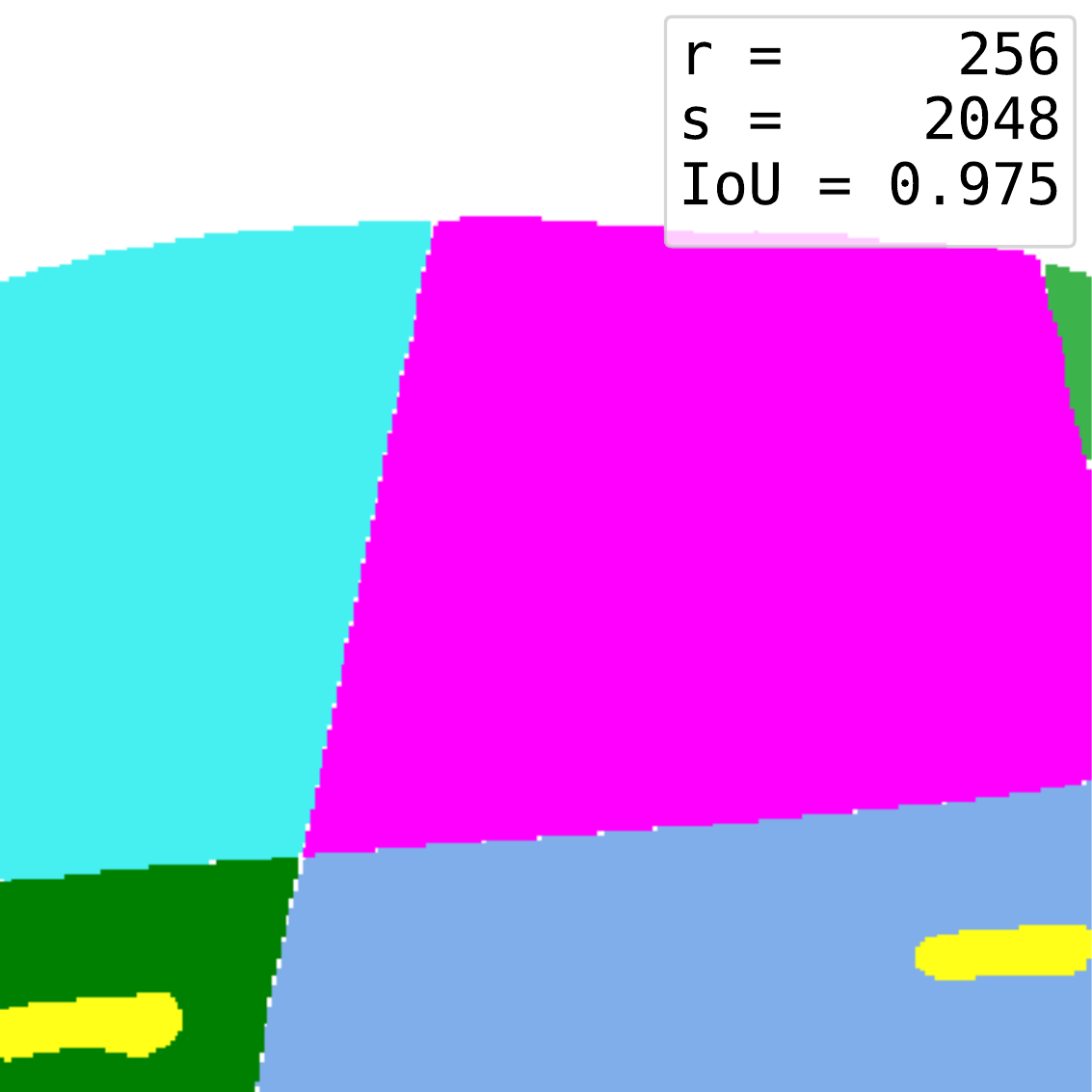}}
    \label{fig:stacked_predictions_2048_synthetic}
    \caption{Segmentation map predictions of U-net models trained with pure real images (top) vs. the same training sets augmented with 2048 synthetic images (bottom).  The input image and ground truth are shown on the left for reference.}
    \label{fig:stacked_predictions}
\end{figure*}


\section{Transfer Learning}
\label{sec:methodology_transfer}

Another potential use case for synthetic data is in pretraining models for later improvement with real data, either as a base for multiple specialized models or as a starting point for incremental training as real data becomes available. 
Our results from the previous section indicate that U-net models trained with 256 or fewer images from our real image dataset suffer from low applicability to new images, so in this section we will focus on pretrained model refinement with small numbers of real images.

The goals and requirements for transfer learning can vary widely, but in our exploration we will focus on use cases stemming from unavailability of real labelled training images and from the need to specialize a general model for a particular task.  As such, we will quantify results in terms of accuracy (in this case, mean prediction IoU on real data) and training time of the model specialization training.

\subsection{U-Net}

There are many strategies for transfer learning using the U-Net model [cite], most involving freezing, reinitializing, adding, or removing layers.
It is beyond the scope of this work to explore the many factors involved in choosing the optimal strategy for a particular use case.
We will instead focus on a relatively simple technique that compares well to our work with a more advanced model in the next subsection, which to train a U-Net with purely synthetic data, and then continuing training with real images while optionally freezing or replacing part of the model.  
Our base synthetic-trained U-Net model uses parameters as described in the previous section, trained with a larger dataset of 36,480 synthetic images, which achieved $0.954$ mean prediction IoU on the holdout set from the same synthetic domain.
Accuracy on segmentation of real images was similar to the experiments with large pure synthetic datasets in the previous section, only achieving a mean prediction IoU of $0.618$ on that domain.

Starting with an identical U-Net base model initialized with random weights, experiments were configured as follows:

\begin{itemize}
    \item \textit{synth-random} - only the contracting path (\textit{encoder}) was initialized with weights from the pretrained base, allowing the untrained expanding path (\textit{decoder}) to train completely on real data;
    \item \textit{synth-synth} - both the encoder and decoder were initialized with pretrained base weights;
    \item \textit{VGG19-random} - the encoder part of the model was replaced with VGG19, detailed below, and the decoder left with random weights;
    \item \textit{VGG19-synth} - the encoder was replaced with VGG19, and the decoder initialized with pretrained base weights;
    \item \textit{control} - the base model was used without freezing or replacing layers, and the initial random weights were unchanged.  Note that this is the same configuration as models in the previous section, and the resulting model is trained on purely real data.
\end{itemize}

Finally, we doubled the above configurations with another parameter, choosing to either freeze the layers of the encoder portion of the model or allow the secondary training with real data to propagate and update the encoder weights.
Our expectations were that freezing the encoder section of the model would reduce training time as there were less parameters to update with each back-propagation, but could reduce the model's ability to adapt to the new data.
Table~\ref{tab:model_size_time} details the number of trainable parameters and mean training time per image-epoch for the four resulting model architectures, which indeed shows decreased time per image with less parameters to update.

For some experiments, the encoder layers of the model were replaced with a VGG19\cite{simonyan_very_2014} model pretrained with weights from ImageNet\cite{deng_imagenet_2009}, following the same procedure as the work done in \cite{jha_doubleu-net_2020} for comparability.
With the models initialized with pretrained weights, we continued training using randomly selected subsets of real images until convergence, using the stopping criteria described in the previous section.  All model variant and real image sample size permutations were repeated 30 times.

\begin{table*}[t]
\centering
\caption{Model Size and Training Time}
\label{tab:model_size_time}
\begin{tabular}{c c c c c c}
    \hline
    & & \multicolumn{2}{c}{parameters (millions)} & training time per \\
    model & variant & total & trainable & epoch-image (s) \\
    \hline
    U-net &          &  7.77 &  7.77 & 0.0130 \\
    U-net & frozen encoder &  7.77 &  3.05 & 0.0120 \\
    U-net & VGG19 encoder         & 23.86 & 23.86 & 0.0176 \\
    U-net & frozen VGG19 encoder & 23.86 &  3.83 & 0.0153 \\
    W-net & frozen 1st U-net & 10.11 &  2.34 & 0.0146 \\
    W-net & frozen VGG19 encoder & 26.59 &  6.56 & 0.0195 \\
    \hline
\end{tabular}
\end{table*}

\renewcommand{\w}{0.93}
\begin{figure}[t]
    \sidesubfloat[]{
        \includegraphics[width=\w\textwidth]{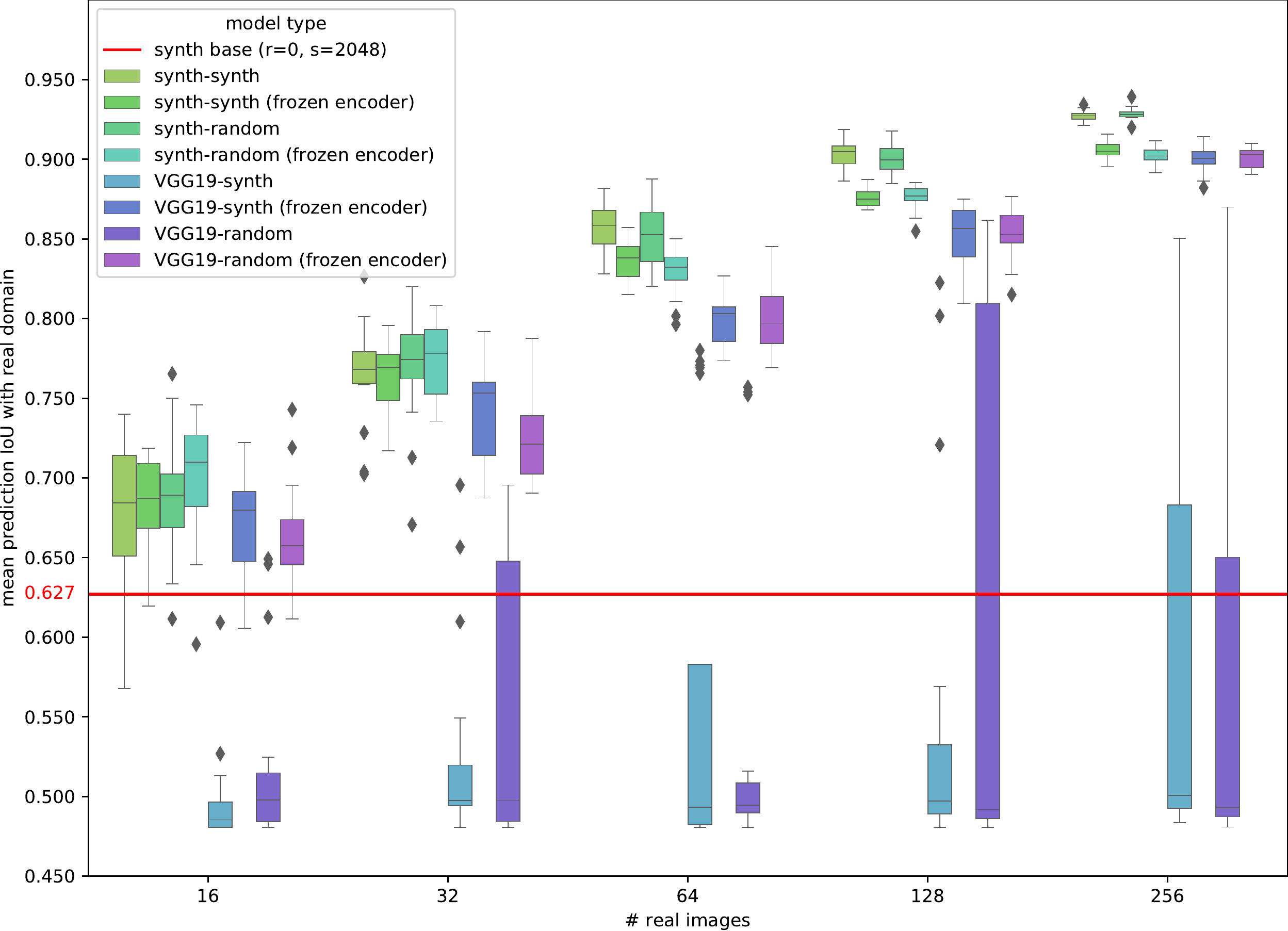}
        \label{fig:transfer_unet_freeze_iou}
    }
    \\
    \vspace{0.5cm}
    \sidesubfloat[]{
        \includegraphics[width=\w\textwidth]{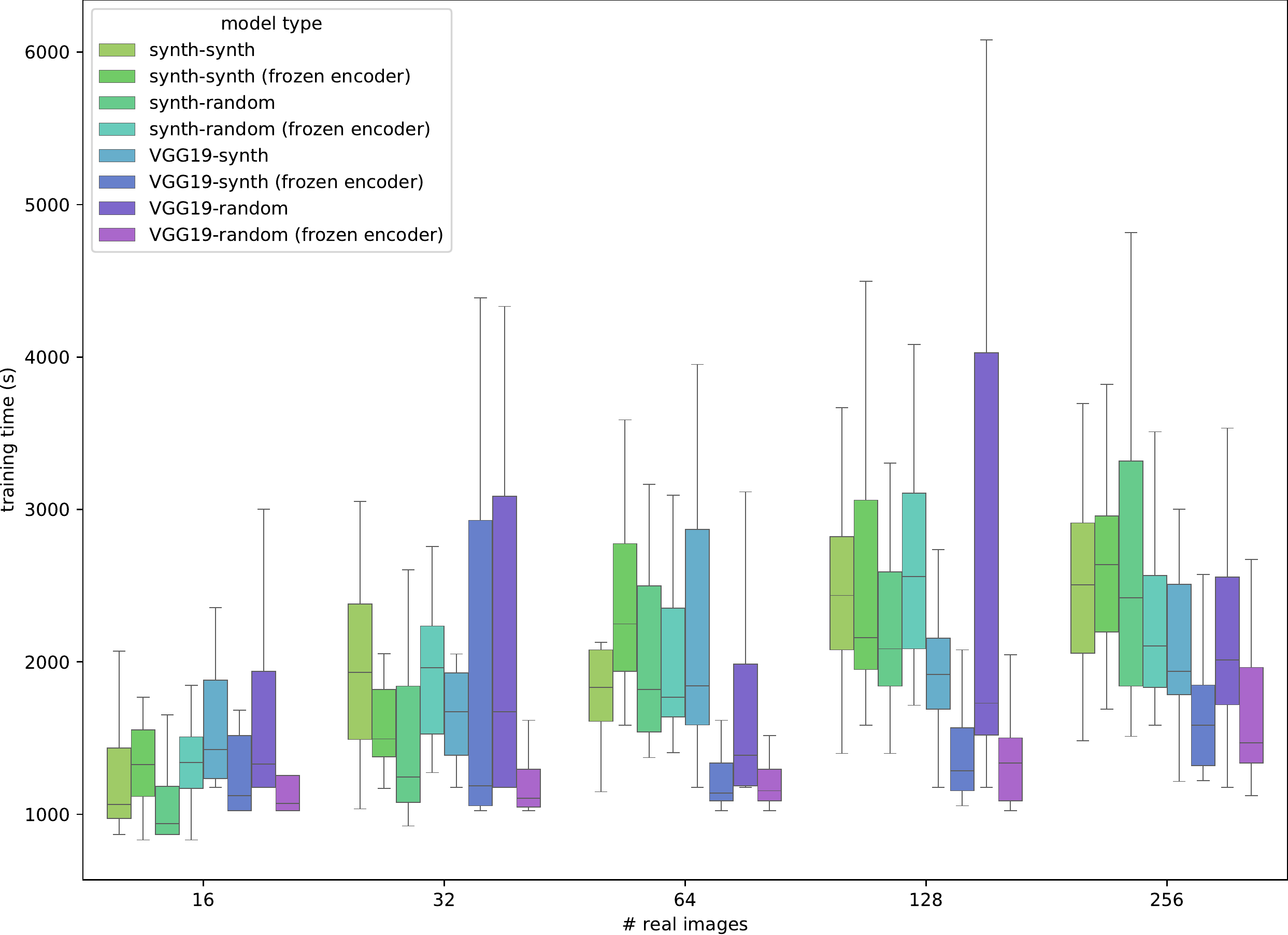}
        \label{fig:transfer_unet_freeze_time}
    }

    \caption{
        Comparisons of mean prediction IoU \protect\subref{fig:transfer_unet_freeze_iou} and training time \protect\subref{fig:transfer_unet_freeze_time} of secondary training of pretrained U-Net models, with the weights of the contracting path (\textit{encoder}) either trainable or frozen.
    }
    \label{fig:transfer_unet_freeze}
\end{figure}

We first compare on the frozen/trainable encoder variable, visualized in Figure~\ref{fig:transfer_unet_freeze}.  In models using VGG19 as the encoder, we observed greater prediction accuracy and lower training time, while models using our encoder pretrained on synthetic data tended to perform better when the encoder was not frozen during secondary training.  This is perhaps due to the large difference in the number of encoder neurons, as propagating the training feedback from each example through the larger VGG19 encoder is more costly and less impactful.
We speculate that limiting the neurons being updated each epoch lead to faster model convergence while the models with more trainable weights slowed in training progress enough to trigger early stopping.
The training logs support this conjecture, showing extremely slow improvement before training was terminated.
It is possible that, given enough time, the accuracy differences between trainable and frozen versions of the same model would minimize.
However, since all models use the same early stopping criteria, we present the results as comparable in a practical sense. 
In the remainder of this work, comparisons with these models will use the better-performing frozen encoders in the case of VGG19, and trainable encoders for the synthetic data-trained models.

\begin{figure}[t]
    \sidesubfloat[]{
        \includegraphics[width=\w\textwidth]{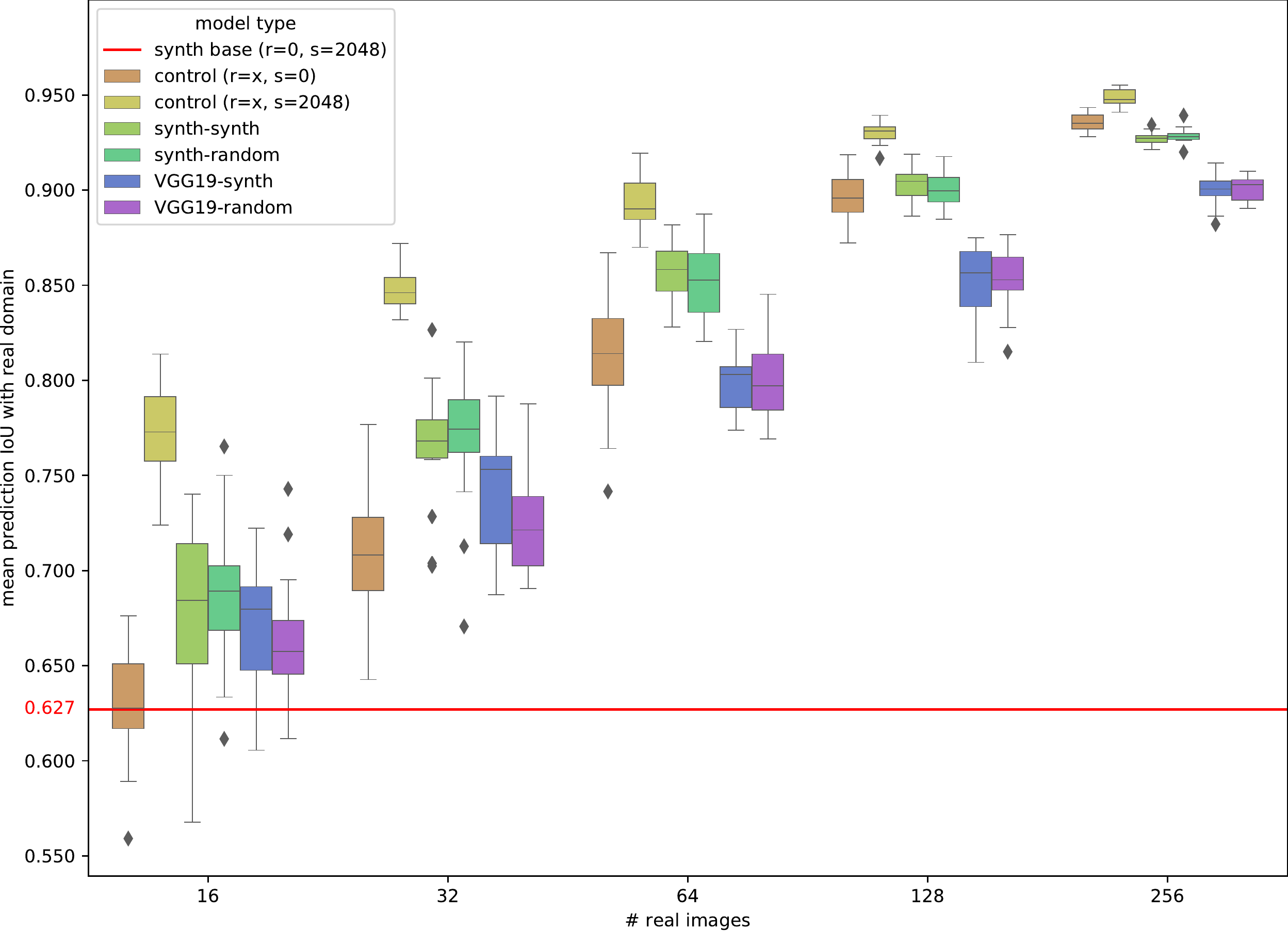}
        \label{fig:transfer_unet_iou}
    }
    \\
    \vspace{0.5cm}
    \sidesubfloat[]{
        \includegraphics[width=\w\textwidth]{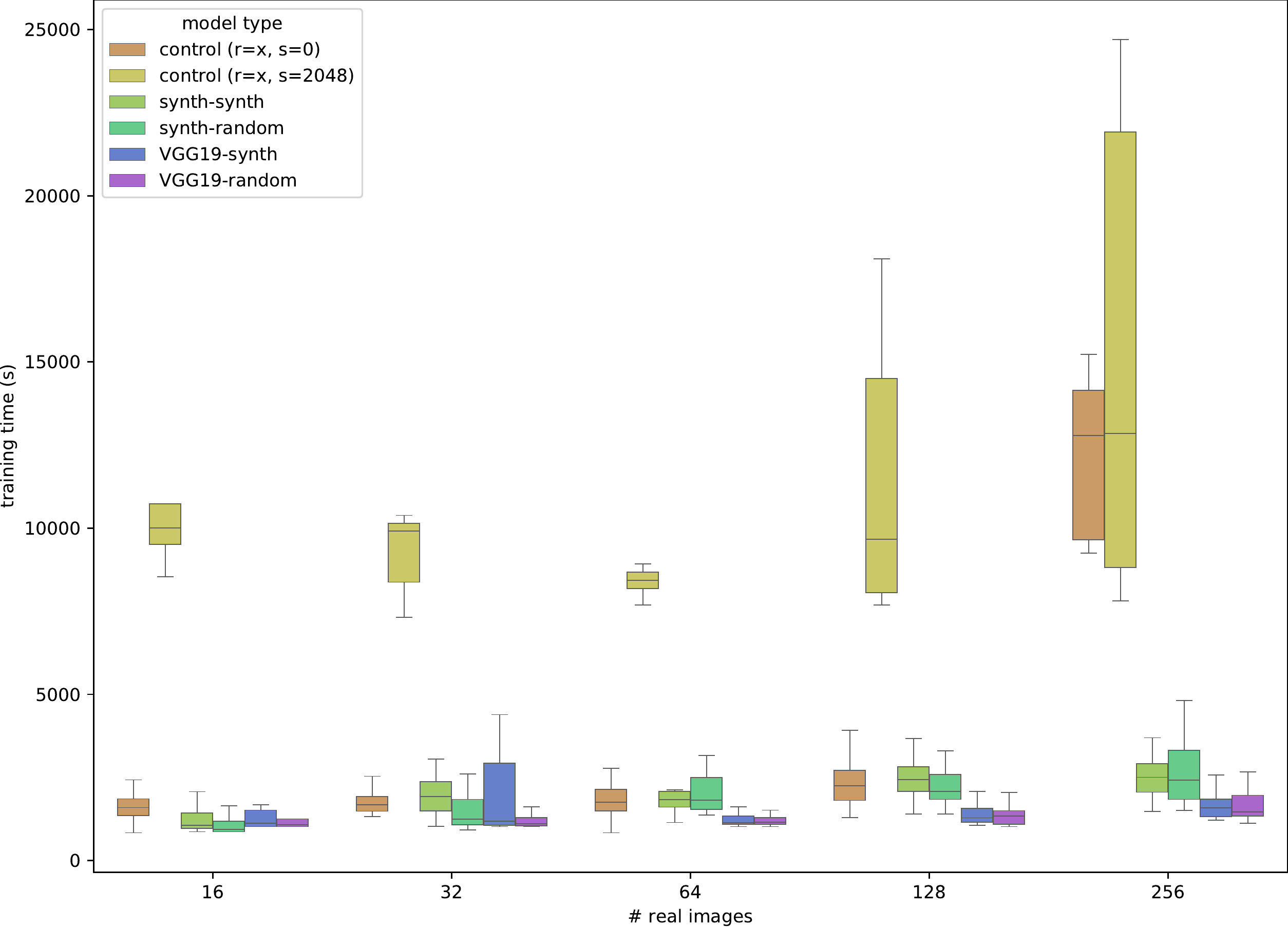}
        \label{fig:transfer_unet_time}
    }

    \caption{
        Limiting to encoder type and decoder initial weights (synthetic pretrained vs. random) model permutations, we observed a sizeable tradeoff between mean prediction IoU \protect\subref{fig:transfer_unet_iou} and training time \protect\subref{fig:transfer_unet_time} when compared to models initialized from randomness.
    }
    \label{fig:transfer_unet}
\end{figure}

Next, we compare the mean prediction accuracy of the retrained models with frozen encoders to the control models trained from randomly initialized weights.  We observed that in cases with 64 or fewer real images, we saw an increase in accuracy over a control model trained on purely real data.  However, in larger real image classes and with all control models trained on a mix of real and synthetic data, we saw significantly lower accuracy in the specialized models.  We again speculate that the model training may have slowed enough to trigger our early termination criteria, and that a combination of refined learning rate, early termination parameters, and lengthened training time may result in improved accuracy.  Our goals in this work are in comparability between experiments, though, so we present these results as a baseline to be improved upon.

In comparing the prediction accuracy of U-net models with different decoder weights, we saw mixed results; the pretrained synthetic data weights appeared to result in lower performance in models with synthetic weighted encoders trained on 16 or 32 real images, while having the opposite effect in models with VGG19 encoders.
In models trained on 64 or more real images, the results were less clear; and a two-sided T-test showed insufficient difference to conclude that the results are drawn from different distributions at $p=0.05$.

Comparing encoder paths of the different model classes was more consistent, in that the U-net default layers trained with synthetic data resulted in higher mean prediction accuracy than models using the VGG19 encoder trained on ImageNet, across all real data sample sizes.
We conclude from these findings that a relatively small encoder (4.72m parameters) trained on a few thousand images drawn from a \textit{similar} synthetic domain to the target can outperform the already impressive feature extraction of a large (23.03m parameters) encoder trained on over a million generic real images.

Figure~\ref{fig:transfer_unet_time} compares the training times of retrained models to those of the control model for each real sample size class, with results between 10.0\% and 20.8\% of the time required for the control.
The time can be accounted for in both the number of trainable parameters in the retrained models with frozen encoders, and the number of epochs required to converge.
As the mean training time for a purely synthetic U-net (r=0, s=2048) is 11,648 seconds, the training time for a retrained U-net is comparable to that of the control.

\subsection{Double-U-Net}

Since the introduction of U-net in 2015, a number of derivative models have been proposed that improve its applicability to certain use cases.
One of these, the Double-U-net\cite{jha_doubleu-net_2020}, improves upon the localization of segment instances by dividing the task between, as the name suggests, two U-net models linked together.
The first U-net, using a VGG19 encoder trained on ImageNet, outputs feature maps from each level of the encoding process as well as an intermediate segmentation map from the decoder.
The segmentation map is paired with the original image as input to the second U-net, while feature map outputs of the first U-net are linked to corresponding layers of the second U-net decoder.
The authors' results showed impressive accuracy gains over a standard U-net on a variety of medical segmentation datasets.

As an exercise in applying transfer learning to a more complex model, we chose the Double-U-net (abbreviated \textit{W-net} for the remainder of this work) because of its intuitive design as a logical extension to the standard U-net, as well as having experience and success using the model in some production use cases.
Our experiments in this section will expand on the previous section for ease of comparison, with the caveat that we made some implementation choices toward this goal while potentially sacrificing some peak performance.
For example, the authors of W-net used \textit{squeeze-excite blocks}\cite{hu2018squeeze} at the end of each convolutional block, which is not part of the original U-net specification. 
Additionally, in our image set, vehicle features were largely scale-invariant, as the images were captured from a fixed viewpoint with a low variation in the vehicle's distance from the camera.  This warranted omission of the Atrous Spatial Pyramid Pooling (ASPP) block between the encoder and decoder in each U-net, which was used in \cite{jha_doubleu-net_2020} to handle feature scaling.
We conducted a limited exploration and found these features to contribute little to no performance gains on our particular use case, so we believe that the simplified model is a better comparison to transfer learning results on a simple U-net in the previous section.


Our W-net implementation is simply two U-net models, identical to the implementation described in the previous section, with the following two additions.  First, as in the \cite{jha_doubleu-net_2020}, the U-nets are connected with a pixel-wise multiplication layer, such that the second U-net receives the original image augmented with the segmentation map output of the first U-net.  Second, the encoder layer-wise feature maps from the first U-net are concatenated to the inputs of the second U-net decoder, in the same manner as the feature maps from the second U-net encoder.  

Following the work in the previous section and as an analog to \cite{jha_doubleu-net_2020}, we chose to construct Double-U-nets with two model variations.  In the first model, we use a U-net trained on synthetic data as described above, with the entire first U-net frozen.  The second model, analogous to \cite{jha_doubleu-net_2020}, uses a frozen VGG19 encoder and a trainable uninitialized decoder.  In both models, the second U-net is initialized with random weights and is fully trainable.
Our hyperparameter search revealed optimal parameters very close to those used to train the individual U-nets, so we opted to keep the original parameters for comparability.

\begin{figure}[t]
    \sidesubfloat[]{
        \includegraphics[width=\w\textwidth]{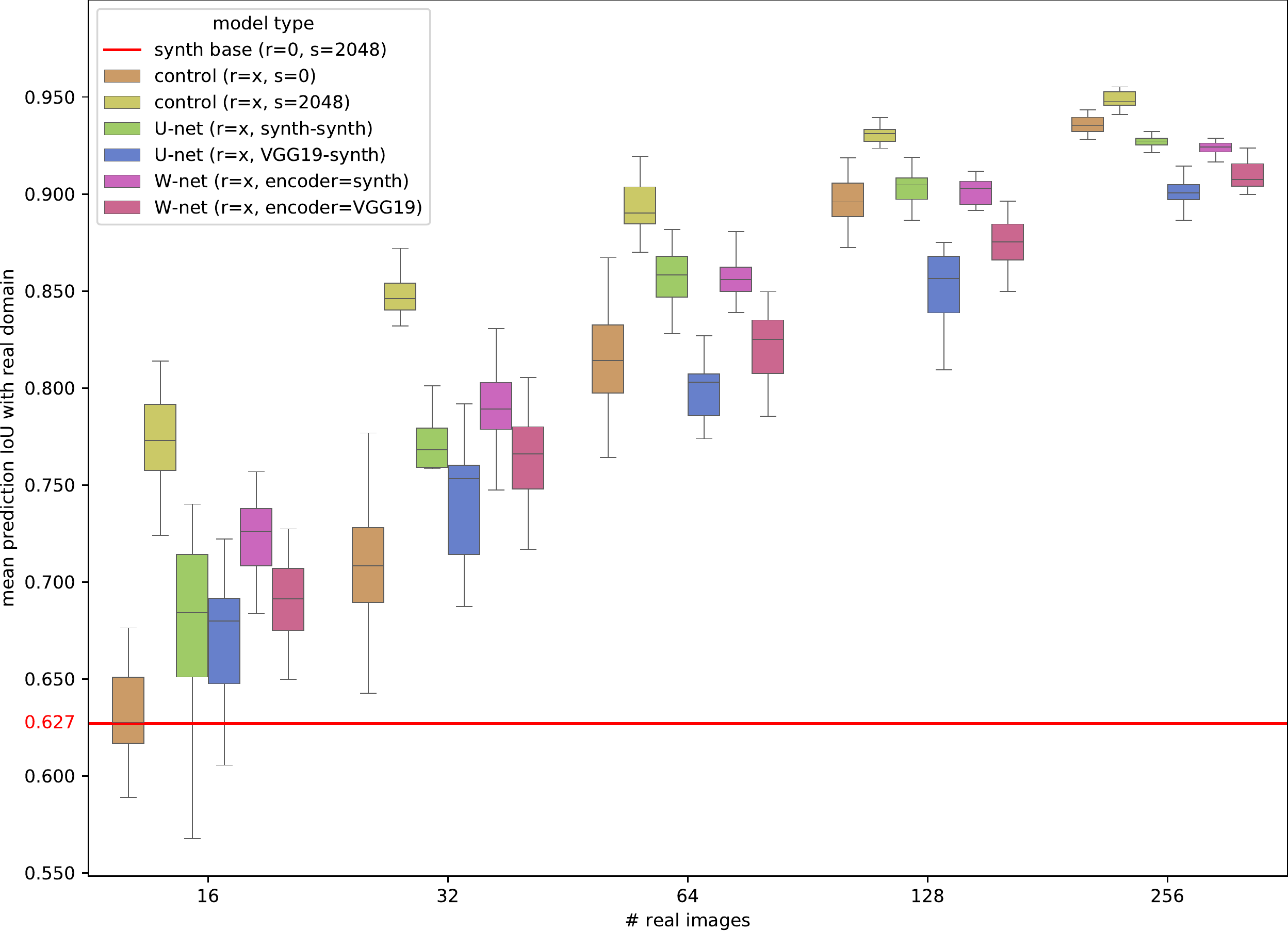}
        \label{fig:transfer_wnet_iou}
    }
    \\
    \vspace{0.5cm}
    \sidesubfloat[]{
        \includegraphics[width=\w\textwidth]{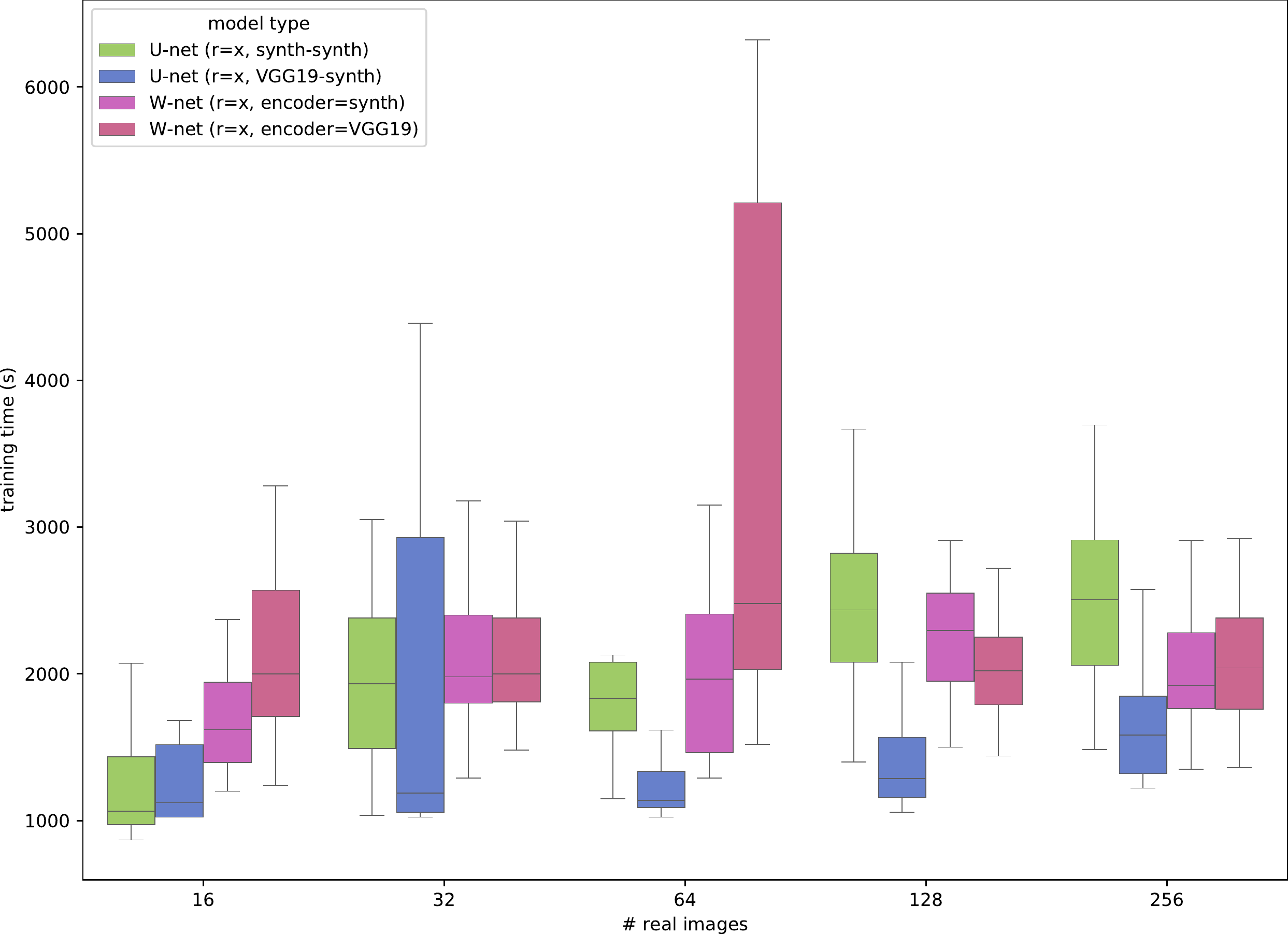}
        \label{fig:transfer_wnet_time}
    }

    \caption{
        Results of transfer learning on U-net and W-net models with (first) encoders trained on synthetic data or VGG19/ImageNet, compared to training of \textit{control} models initialized with random weights.
        The mean prediction IoU \protect\subref{fig:transfer_wnet_iou} and training time \protect\subref{fig:transfer_wnet_time} suggest improved accuracy of synthetic-trained encoders, but in some cases with a time cost.
    }
    \label{fig:transfer_wnet}
\end{figure}

Our results, shown in Figure~\ref{fig:transfer_wnet}, show accuracy improvements using the W-net model with the VGG19 encoder over all training image size classes, and similar or better results with the synthetic-trained first U-net.
The accuracy improvements correlate with a training cost increase, however, especially with the VGG19-based models with more layers to train.  
The conclusion we draw from these results is that secondary training with a multipart model like W-net can be a viable accuracy enhancement if the time cost can be justified.

\section{Conclusions}
\label{sec:conclusions}
We found that, for this image segmentation problem, synthetic images were an effective technique for augmenting limited sets of real training data.  
We observed that models trained on purely synthetic images had a very low mean prediction IoU on real validation images.
We also observed that adding even very small amounts of real images to a synthetic dataset greatly improved accuracy, and that models trained on datasets augmented with synthetic images were more accurate than those trained on real images alone.
We noted that for this domain, 256 to 512 images seemed to be enough to train a reasonably accurate model, with rapidly diminishing returns on adding synthetic images to the mix, eventually resulting in lower accuracy as the real:synthetic ratio dropped.

In use cases that benefit from incremental training or model specialization, we found that pretraining on synthetic images provided a usable base model for transfer learning.
While we observed that models trained in a single session outperformed those pretrained on synthetic images and retrained on real data, we also saw that up to 90\% of the total training time could be completed in the pretraining phase.

We conclude that synthetic image generation can be beneficial to segmentation model training when insufficient images are available to train a satisfactory model.
However, testing must be done to find the break point where adding more synthetic images does not result in higher mean accuracy.


\section{Future Work}
\label{sec:future}
A natural progression from this work is to study the characteristics of synthetic data and identify features that contribute to model accuracy and can be adapted to more closely resemble the real domain, while separating less important features that should be randomized.
Recent work in the field of Generative Adversarial Networks (GANs) could be used to automate the feature identification process and help design more robust synthetic image rendering processes.
Another interesting topic would be exploring how synthetic images can be used in conjunction with other effective data augmentation techniques, which unfortunately was beyond the scope of this work.



\newpage

\bibliographystyle{IEEEtran}
\raggedright
\bibliography{references}

\end{document}